\documentclass{article}

\PassOptionsToPackage{numbers,sort&compress}{natbib}

\usepackage[dblblindworkshop,final]{neurips_2026}
\workshoptitle{Managing Agents that Manage Agents:
  Workshop on Responsible Use of Meta-Agents}
\usepackage[utf8]{inputenc}
\usepackage[T1]{fontenc}
\usepackage{microtype}
\usepackage{url}
\usepackage{graphicx}
\usepackage{multirow}
\usepackage{comment}
\usepackage{algorithm}
\usepackage{algorithmic}
\usepackage{float}
\usepackage{subcaption}      
\usepackage{array}
\usepackage[export]{adjustbox}
\usepackage{needspace}
\usepackage{amsmath,amssymb}
\usepackage{placeins}
\usepackage{xcolor}
\usepackage{enumitem}
\usepackage{caption}
\usepackage{hyperref}
\usepackage{xcolor}

\setlist{
  topsep=0.35em,
  itemsep=0.15em,
  parsep=0pt,
  partopsep=0pt
}

\makeatletter

\@ifpackageloaded{array}{}{\usepackage{array}}
\@ifpackageloaded{longtable}{}{\usepackage{longtable}}
\@ifpackageloaded{booktabs}{}{\usepackage{booktabs}}
\@ifpackageloaded{xcolor}{}{\usepackage{xcolor}}
\@ifpackageloaded{colortbl}{}{\usepackage{colortbl}}
\@ifpackageloaded{amssymb}{}{\usepackage{amssymb}}
\@ifpackageloaded{textcomp}{}{\usepackage{textcomp}}

\DeclareTextCommand{\textquotedbl}{OT1}{\textquotedblright}
\DeclareTextCommand{\guillemotleft}{OT1}{\ensuremath{\ll}}
\DeclareTextCommand{\guillemotright}{OT1}{\ensuremath{\gg}}

\definecolor{GEPAname}{RGB}{16,78,139}
\definecolor{GEPAband}{RGB}{225,236,248}
\definecolor{NaiveName}{RGB}{166,63,0}
\definecolor{NaiveBand}{RGB}{252,236,219}
\definecolor{GameBand}{RGB}{228,228,228}
\definecolor{SelGray}{RGB}{105,105,105}

\@ifundefined{PromptColW}{\newlength{\PromptColW}}{}
\@ifundefined{PromptBandW}{\newlength{\PromptBandW}}{}
\providecommand{\PromptBandSize}{\small}
\providecommand{\PromptAnnSize}{\tiny}

\def\PromptBand#1#2#3#4{%
  \rowcolor{#1}\multicolumn{2}{>{\raggedright\arraybackslash}p{\PromptBandW}}{%
    \PromptBandSize\bfseries\textcolor{#2}{#3}%
    \ifx\relax#4\relax\else
      \newline\normalfont\PromptAnnSize\textcolor{SelGray}{#4}%
    \fi}%
}

\makeatother

\title{Naive Prompt Optimization: \\
Rethinking the Need for Complex Prompt Search}

\workshoptitle{Managing Agents that Manage Agents: Workshop on Responsible Use of Meta-Agents}

\author{%
  Yuan Chang \qquad Xiaoqi Chen \\
  School of Electrical and Computer Engineering, Purdue University \\
  \texttt{\{xiaoqic,chang658\}@purdue.edu}
}

\begin{document}

\maketitle

\begin{abstract}

Efficiently improving autonomous agents across diverse tasks is central to accelerating recursive self-improvement (RSI) in agentic AI, with prompt optimization emerging as a promising approach capable of delivering performance gains comparable to those achieved by fine-tuning model weights, while reducing computational costs in both optimization and serving.
However, recent developments increasingly favor unnecessarily complex prompt optimizers. 
We introduce Naive Prompt Optimization (NPO), a lightweight single-lineage method that iteratively revises prompts using a teacher model with rollout feedback. NPO achieves comparable or better performance than GEPA with fewer rollouts in IFBench and HotpotQA, and its advantage increases with stronger teacher models, suggesting that stronger teacher reasoning may partially substitute for optimizer-side search complexity. In interactive games, NPO remains broadly competitive with GEPA despite its much simpler mechanism, while GRPO still performs better on some tasks less amenable to prompt-only optimization. 
We also show that NPO-optimized prompts elicit similar performance improvements when applied verbatim to other student models, especially across models within the same family. Overall, our preliminary results show that simple, linear prompt optimization can rival substantially more sophisticated and complex search procedures.
\end{abstract}

\section{Introduction}

Reinforcement learning (RL) and automatic prompt optimization offer two complementary ways to improve LLM performance. RL methods, including PPO~\citep{ppo}, RLHF~\citep{instructgpt}, and GRPO~\citep{deepseekmath}, update model parameters from reward signals, while prompt optimization keeps model parameters fixed and instead improves the instructions guiding model behavior,
enabling lightweight, portable adaptation to individual users and tasks at deployment time through standard third-party inference APIs, without the cost and complexity of maintaining and serving user- or task-specific model weights.

Recent prompt optimization methods increasingly use sophisticated search strategies. OPRO~\citep{opro} treats an LLM as an optimizer that proposes new instructions from previously evaluated solutions and scores.  
AI4AI~\citep{ai4ai} studied how a teacher model builds an inference-time harness for weaker student models. ProTeGi~\citep{protegi} combines textual gradients with beam search, MIPRO~\citep{mipro} uses model-generated proposals and Bayesian optimization, and GEPA~\citep{gepa} maintains multiple candidates through reflection and Pareto-based selection.

Against this background, we introduce Naive Prompt Optimization (NPO), a lightweight single-lineage method that iteratively revises the prompt by providing rollout traces and rewards to a teacher model (also referred to as a reviser or reflection model), which generates the next prompt version without maintaining multiple prompt lineages or using explicit search algorithms. We then systematically compare different prompt optimization methods and GRPO across established benchmarks~\citep{ifbench,hotpotqa} and interactive game environments~\citep{textarena}.

Across these settings, we vary the teacher model used for prompt revision and find that NPO’s advantage over GEPA widens as the teacher model becomes stronger. We also vary the student model and demonstrate that prompts optimized on one student model can be applied directly to other models to show similar performance gains.

Our main contributions are:

\begin{enumerate}

\item We propose Naive Prompt Optimization (NPO), a simple LLM-as-optimizer method that iteratively revises prompts for a fixed student LLM using full rollout trajectories and rich feedback, rather than only previous prompts and scores as in OPRO.

\item We develop a controlled, low-variance methodology for isolating improvements in decision-making performance from prompt optimization and reinforcement learning, using shared pseudorandomness in environment generation to enable fair comparisons and constrained decoding to eliminate formatting-induced penalties.

\item We show prompts resulting from NPO bring transferable improvements, yielding similar performance gains when applied verbatim to other student models within and across families.

\item{
We evaluate the impact of prompt optimization and reinforcement learning on speculative decoding efficiency, particularly the acceptance rates and inference speed of pretrained multi-token prediction (MTP) models in different tasks.
}

\end{enumerate}

Empirically, NPO achieves better performance
under the same rollout budgets as the substantially more complex GEPA prompt search procedure on IFBench, which evaluates instruction following under verifiable constraints~\citep{ifbench}, and HotpotQA, a multi-hop question-answering benchmark~\citep{hotpotqa,bm25,rag}. 
NPO's advantage is particularly pronounced with stronger teacher models, suggesting that stronger teachers combined with rich feedback may reduce the need for complex prompt-search procedures. Across 22 TextArena games involving complex strategic planning and decision-making, NPO exhibits broadly comparable performance to GEPA despite its much simpler mechanism, while GRPO provides complementary gains on several tasks where prompt optimization is less effective. 
The performance gains observed in NPO also generally transfer across model scales and families, with only minor variation, when the optimized prompts are directly applied unmodified on different student models, especially for students from the same model family.

\begin{figure}[H]
    \centering
    \vspace{-0.5em}   
    \includegraphics[width=0.99\textwidth]{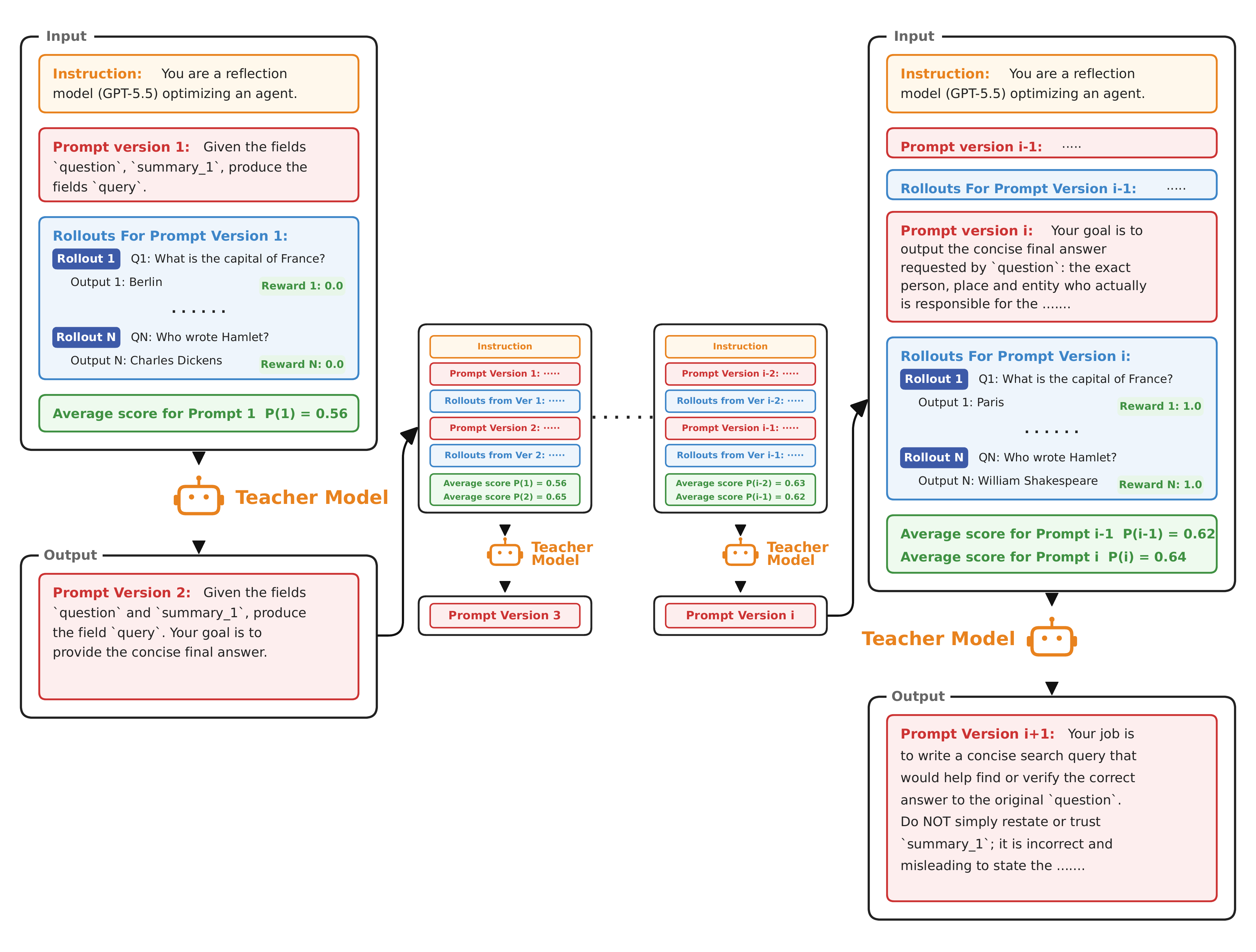}
    \vspace{-1em}
    \caption{Overview of the Naive Prompt Optimization (NPO) workflow.}
    \label{fig:overview}
\end{figure}

\section{Methodologies and experiment setup}
\label{sec:methods}

We first introduce our proposed method, Naive Prompt Optimization (NPO), and then briefly review the established GEPA and GRPO methods, together with our  experimental setup.

\subsection{Naive Prompt Optimization (NPO)}
\label{sec:naive}

Naive Prompt Optimization maintains a single prompt lineage and iteratively updates it using execution feedback. 
As illustrated in Figure~\ref{fig:overview}, at iteration $i$, we execute the student model with the current version of the prompt $\mathcal{P}^{(i)}$, repeating $N$ times for a minibatch of rewards and \mbox{rollout traces~$\mathcal{R}_i$.}
We then feed the teacher model a sliding window of the $W$ most recent iterations---from version $(i-W+1)$ through $i$---including the prompts, corresponding rollout traces, and rewards. Using this rich context, the teacher revises the previous prompts to produce the next prompt, $\mathcal{P}^{(i+1)}$.

NPO uses an LLM as an iterative prompt optimizer (teacher), which was first proposed in OPRO~\citep{opro}; unlike OPRO, which conditions optimization only on previously evaluated prompts and their scalar scores, NPO revises prompts using complete rollout traces and rollout-specific rewards.

\begin{algorithm}[t]
\caption{Naive Prompt Optimization with Sliding-Window Rollout Feedback}
\label{alg:naive}
\footnotesize
\begin{algorithmic}[1]

\REQUIRE Initial prompt $\mathcal{P}^{(0)}$, sample task dataset $\mathcal{D}$ from environment,
minibatch size $N$, window size $W$, teacher $\mathcal{T}$, iterations $Y$

\FOR{$i=0$ \TO $Y-1$}
    \STATE Sample minibatch $\mathcal{B}_i$ of size $N$ from $\mathcal{D}$
    \STATE Run the target model with prompt $\mathcal{P}^{(i)}$ on $\mathcal{B}_i$
    \STATE Collect rollout traces and rewards $\mathcal{R}_i$

    \STATE Construct sliding-window feedback:
    \[
    \{\mathcal{R}_j\}_{j=\max(0,i-W+1)}^{i}
    \]
    
    \STATE Generate the next prompt:
    \[
    \mathcal{P}^{(i+1)}
    \leftarrow
    \mathcal{T}\!\left(
    \mathcal{P}^{(i)},
    \{\mathcal{R}_j\}_{j=\max(0,i-W+1)}^{i}
    \right)
    \]
\ENDFOR

\RETURN Prompt sequence
$\{\mathcal{P}^{(0)},\ldots,\mathcal{P}^{(Y)}\}$ and best candidate

\end{algorithmic}
\end{algorithm}

\subsection{GEneric-PAreto (GEPA)}
\label{sec:gepa}
\begin{figure}[t]
    \centering
    \includegraphics[width=0.88\textwidth]{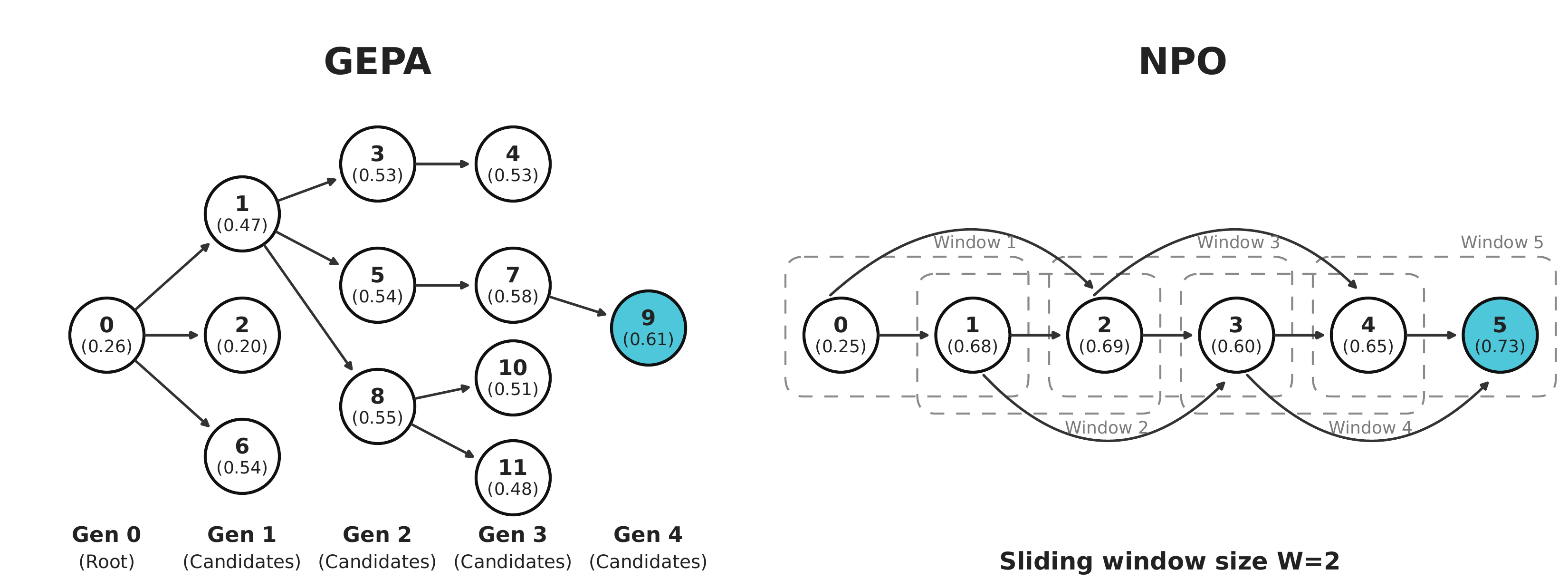}
    \caption{GEPA evolves candidate prompts in a pool, while NPO follows a simple lineage.}
    \label{fig:npo-vs-gepa}
\end{figure}

GEPA~\citep{gepa} maintains a pool of prompt candidates and expands it through reflection and Pareto-based selection. A parent prompt is sampled from the candidate pool, revised using feedback from a reflection minibatch, and retained only if the revision improves on that minibatch. Accepted revisions are then evaluated on the validation set and added to the candidate pool, where instance-level validation performance informs subsequent Pareto-based parent selection. Figure~\ref{fig:npo-vs-gepa} illustrates GEPA's prompt heritage tree, compared to NPO's linear evolution.

We use the standard GEPA procedure without the GEPA-merge variant. For IFBench and HotpotQA, we follow the original GEPA algorithm directly. For TextArena, we make only a minor adaptation to accommodate interactive game episodes: each reflection example corresponds to a complete target-player game trajectory together with its episode-level reward, while the underlying candidate selection, reflection, acceptance, and Pareto-based selection mechanisms remain unchanged.

\subsection{Group Relative Policy Optimization (GRPO)}
\label{sec:grpo}

We use Group Relative Policy Optimization (GRPO)~\citep{deepseekmath} as a weight-based fine-tuning baseline for comparison with prompt optimization. For each rollout group, GRPO centers the rollout rewards by subtracting their group mean and rescales them by their standard deviation to obtain relative advantages for policy updates. We hold the default game prompt and backbone weights fixed while training only a task-specific low-rank adapter (LoRA)~\citep{lora}.

In two-player environments, the target policy plays against a fixed opponent instantiated from the unmodified base model; only the target policy's LoRA parameters are updated. Because many such games yield different reward distributions depending on whether a player moves first or second, during environment randomization we balance the target policy's move order within every rollout group: it moves first in exactly half of the randomized episodes and second in the remaining half.

\newcommand{\examplebox}[2]{%
\begin{minipage}[t]{0.475\linewidth}
\centering
{\scriptsize\bfseries #1}\\[1pt]
\setlength{\fboxsep}{4pt}%
\fcolorbox{black!20}{gray!5}{%
\parbox[t]{\dimexpr\linewidth-2\fboxsep-2\fboxrule\relax}{%
\raggedright
\scriptsize #2%
}}%
\end{minipage}%
}

\subsection{Experiment setup}
\label{sec:textarena-setup}

\begin{figure}[t]
    \centering
\includegraphics[width=1.0\textwidth]{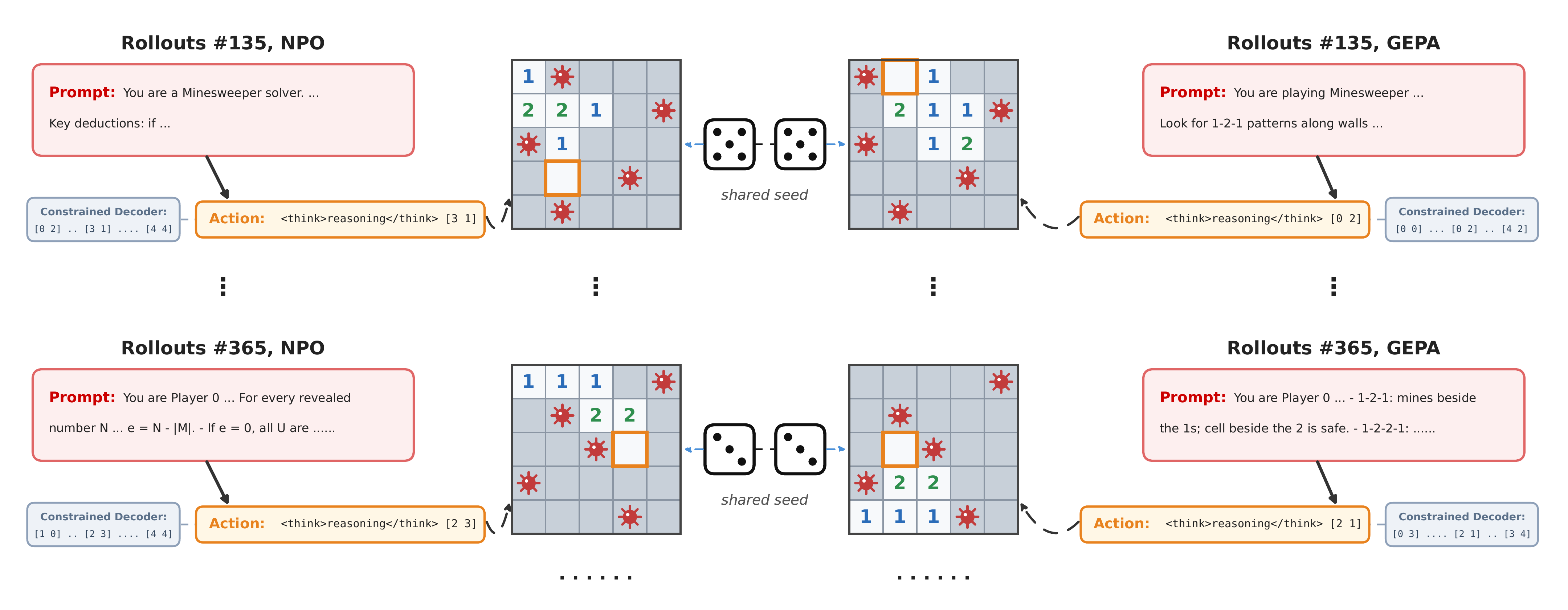}
\vspace{-1em}
    \caption{We use shared pseudo-randomness to reduce environment-induced noise when comparing.}
    \label{fig:pseudorandomness}
\end{figure}

To enable controlled, low-variance comparisons across optimization methods, we carefully design and instrument the evaluation pipeline to minimize sources of variation unrelated to the methods themselves. In particular, our setup controls randomness in environment generation and separates improvements in decision-making from artifacts introduced by output formatting. These designs allow performance differences to be attributed more directly to the optimization method being evaluated.

When generating rollout traces, we use shared pseudorandomness to evaluate all methods on matched environment instances. 
For each environment, we generate a pool of instances from predefined random seeds and reuse the same seed sequence across optimization methods, as illustrated in Figure~\ref{fig:pseudorandomness}. Corresponding rollouts therefore begin from the same randomly generated configuration, even though their actions and subsequent trajectories may differ. This paired evaluation design reduces noise arising from variation in instance difficulty and enables fairer comparisons between optimization methods.

When decoding actions, we use constrained generation to isolate decision-making performance from formatting noise. Our preliminary experiments showed that formatting errors accounted for a nontrivial fraction of observed failures and could therefore distort the comparison between optimization methods. 
We instead require each LLM response to exactly follow the structure \texttt{\textless think\textgreater{} reasoning \textless/think\textgreater{} [action]}.
After the initial prompt and environmental observation,
we prefill the response with the opening \texttt{\textless think\textgreater{}} token, allowing the model to begin generating its reasoning trace immediately.\footnote{Although the student model might not have been specifically fine-tuned for reasoning, the teacher may still elicit chain-of-thought reasoning through few-shot or zero-shot prompting~\citep{wei2022chain,kojima2022large}.} Once the model emits \texttt{\textless/think\textgreater{}}, the decoder switches from unconstrained reasoning to constrained action generation. For environments that expose a finite list of legitimate actions, we compile this environment-provided set of strings into token-level decoding constraints~\citep{lmformatenforcer,vllm}, requiring the model to choose among the actions allowed under the current environment state.

\par\vspace{3pt}
\noindent
\examplebox{Invalid Format under Free Form Decoding}{%
\texttt{\textcolor{blue!70!black}{
\textless think\textgreater{} I think I should go to row 3 column 4
\textless/think\textgreater{} row 3 and then column 4}}
}%
$\Rightarrow$
\examplebox{Enforced Choice from the Legitimate Action Set}{%
\texttt{\textcolor{blue!70!black}{
\textless think\textgreater{} I think I should go to row 3 column 4
\textless/think\textgreater{}}
\textcolor{red!70!black}{[3,4]}}
}%

\par\vspace{5pt}

If the model reaches its allotted reasoning budget without closing the reasoning segment using  \texttt{\textless/think\textgreater{}}, we do not mechanically count the missing action as a failure, which would conflate decision quality with awareness of the token limit. Instead, the decoder inserts the closing token right before the reasoning-budget cutoff  and uses a small reserved action budget to require the model to choose among the legitimate actions.

\par\vspace{3pt}
\noindent
\examplebox{Reasoning Budget Exhausted}{%
\texttt{\textcolor{blue!70!black}{
\textless think\textgreater{} I think I should go to row 3 column 4,
but I need to verify \ldots{} (too long) \ldots{} [limit reached]}}
}%
$\Rightarrow$
\examplebox{Enforced Choice After Cutoff}{%
\texttt{\textcolor{blue!70!black}{
\textless think\textgreater{} I think I should go to row 3 column 4,
but I need to verify \ldots{} (too long) \ldots{}
\textless/think\textgreater{}}
\textcolor{red!70!black}{[3,4]}}
}%

\par\vspace{2pt}

Together, these decoding-time designs isolate improvements in decision-making performance from noise introduced by the random variation in seeding and format-following, while preserving the model’s freedom to generate unconstrained reasoning. 
Details of the constrained-decoding procedure, including the estimation of probabilities over legitimate actions, are provided in Appendix~\ref{app:constrained-decoding-design}.

\subsection{Teacher variation and cross-student prompt transfer}

Our experimental design varies both the teacher model used for prompt revision and the student model executing the optimized prompt. To investigate how teacher capability affects prompt optimization, we hold the student fixed as Qwen3-8B and begin with Qwen3-8B also serving as the teacher, following the same self-revision setting used in GEPA~\citep{gepa}, and then replace it with progressively stronger teachers: DeepSeek-V4-Flash-preview-0424~\citep{deepseekv4} and GPT-5.5~\citep{gpt55}. 
All other components of the setup remain unchanged, allowing us to directly compare the effects of different teacher models.

Besides rerunning prompt optimization separately for every student model, we also investigate the more practically useful scenario of reusing the same prompts optimized on one student directly on other students, without re-running prompt optimization.
For prompts optimized on Qwen3-8B, we evaluate transfer to larger models from the same family, Qwen3-14B and Qwen3-32B~\citep{qwen3}, as well as models from different families, Llama-3.1-70B-Instruct~\citep{llama31} and Llama-3.3-70B-Instruct~\citep{llama33}. We repeat this experiment for prompts optimized on Llama-3.1-8B as the  student, transferring to Llama-3.1-70B-Instruct and Llama-3.3-70B-Instruct models from the same family and Qwen3-32B and StepFun-3.7-Flash~\cite{stepfun2026step37flash} models across families. This design directly tests whether prompt improvements transfer across model scales and families without requiring student-specific re-optimization.

\subsection{Optimization budgets and iteration settings}
\label{sec:iteration_settings}

For IFBench and HotpotQA, we tune NPO's reflection minibatch and sliding-window sizes to provide rich feedback while remaining within the teacher model's context window. NPO uses larger reflection minibatches than GEPA's size of 3~\citep{gepa}, which was selected in the original GEPA study for GEPA's own optimization procedure, yielding minibatch/iteration settings of 50/10 for IFBench and 40/20 for HotpotQA. Each revised prompt is evaluated on a separate 300-example validation set. The resulting NPO rollout budgets are 3,500 and 6,800, slightly below GEPA's 3,593 and 6,871, respectively~\citep{gepa}. Despite larger reflection minibatches, NPO consistently reaches comparable or higher peak performance with fewer total rollouts.
For the 22 TextArena games, we allow NPO and GEPA to use the same rollout budget of 408 total episodes. NPO runs 24 episodes per iteration for 17 iterations, while GEPA uses 6 pseudo-reflection and 18 pseudo-validation episodes per cycle. Because TextArena episodes contain substantially longer interaction trajectories, these settings are chosen to allow rollout history to fit within the teacher model's context window limit (1 million tokens for both GPT-5.5 and DeepSeek-V4-Flash-preview-0424).
We also analyze the sensitivity of NPO and GEPA to  different batch sizes in Section~\ref{sec:batch-window}.

For GRPO, we train a LoRA adapter~\citep{lora} on Qwen3-8B while keeping the backbone and prompt fixed. Based on group-size experiments from 6 to 48, we use 8--12 episodes per iteration for approximately 100 iterations, yielding 800--1,200 training rollouts per game in addition to separate evaluation rollouts. Although we continued GRPO training beyond 408 rollouts for analyzing the effect of batch size, we use the checkpoint at 408 rollouts for comparison against NPO and GEPA under the same rollout budget. 
Episode-level rewards are used to construct the GRPO training signal, with the specific reward formulation depending on the game. GRPO is implemented with DSPy's \texttt{dspy.GRPO} interface~\citep{arborgrpo}, the Arbor RL backend~\citep{arbor}, and ZeRO-3~\citep{zero} on  two NVIDIA H100 GPUs. Additional group-size analysis is provided in Appendix~\ref{app:grpo-group-size}.

For output generation, we use a fixed 2,000-token context window covering the prompt, current observation, and model response, balancing reasoning capacity with training efficiency and computational cost. 
In practice, we observed student models typically use only 800--1200 reasoning tokens and do not exhaust the available budget. Further, we verified that forcing them to continue reasoning beyond their natural termination point until the remaining budget is exhausted does not improve performance; we therefore treat the context window as an upper bound and allow reasoning to terminate naturally.


\section{Results}
\begin{figure}[H]
    \centering
    \vspace{-0.8em}
    {\includegraphics[width=1.0\textwidth]{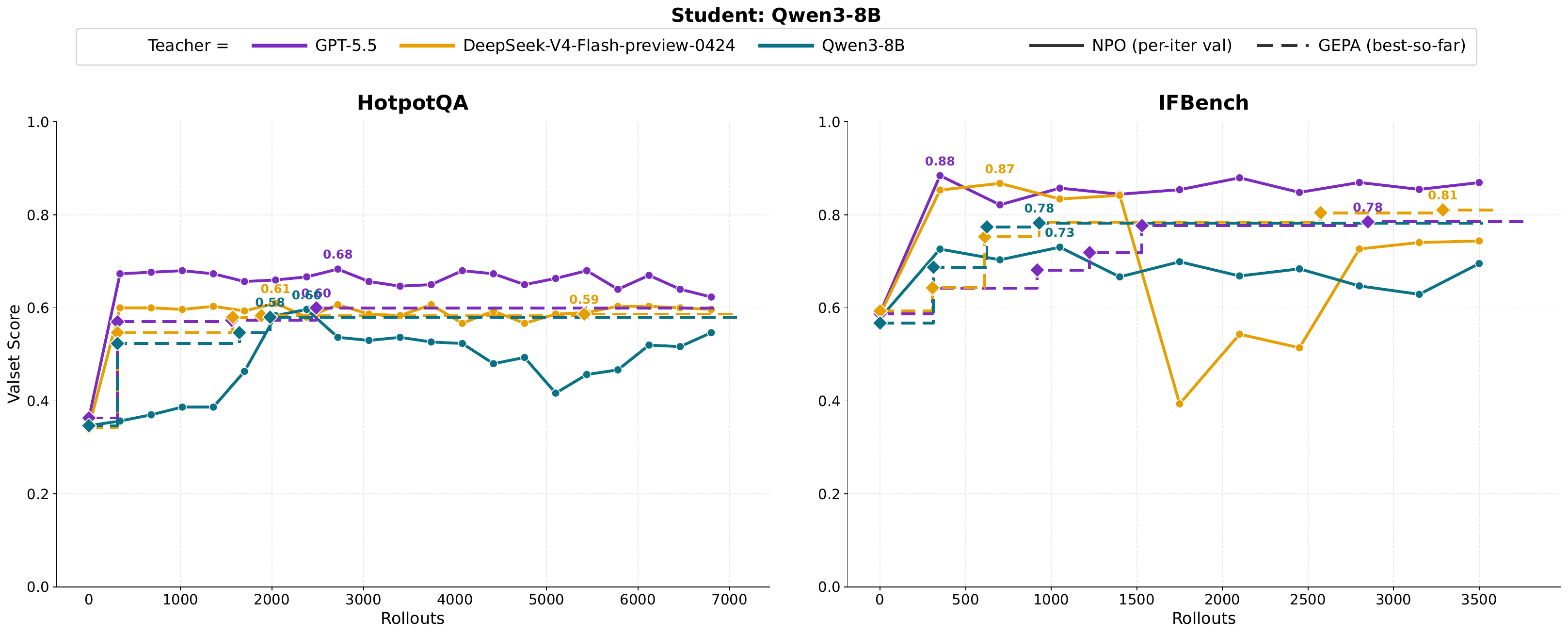}}
    \caption{Performance of NPO and GEPA on IFBench and HotpotQA under different teacher models, showing NPO's advantage with stronger teachers.}
    \label{fig:npo-vs-gepa-teacher}
\end{figure}
\subsection{
NPO outperforms GEPA at the same rollout budgets, given strong teachers}

Figure~\ref{fig:npo-vs-gepa-teacher} compares NPO and GEPA on IFBench and HotpotQA using Qwen3-8B as the fixed student and Qwen3-8B, DeepSeek-V4-Flash-preview-0424, and GPT-5.5 as teachers, with lines color-coded to refer to different teacher models.
As we can see, NPO benefits more consistently from stronger teachers: NPO with GPT-5.5 achieves the fastest convergence and highest validation performance on both tasks. 
In particular, NPO can achieve the same improved performance using fewer rollouts than GEPA.
In contrast, GEPA shows smaller gains from stronger teachers, with GEPA+GPT-5.5 performing broadly on par with GEPA+Qwen3-8B in several settings. These results suggest that stronger teacher reasoning, combined with richer context feedback, may diminish the benefits of more sophisticated prompt-search strategies, allowing a simple iterative method such as NPO to achieve comparable or better performance given the same rollout budget.

\subsection{Optimized prompts are transferable across student models}

One important practical advantage of prompt optimization over weight-based reinforcement learning is portability: an optimized prompt can, in principle, be applied directly to other models, whereas weight-based fine-tuning produces model-specific parameter updates. We now test whether this advantage holds empirically by applying prompts optimized on Qwen3-8B and Llama-3.1-8B unchanged to larger student models, both within and across model families, without re-optimization.
\begin{figure}[t]
    \centering
    \includegraphics[width=0.95\textwidth]{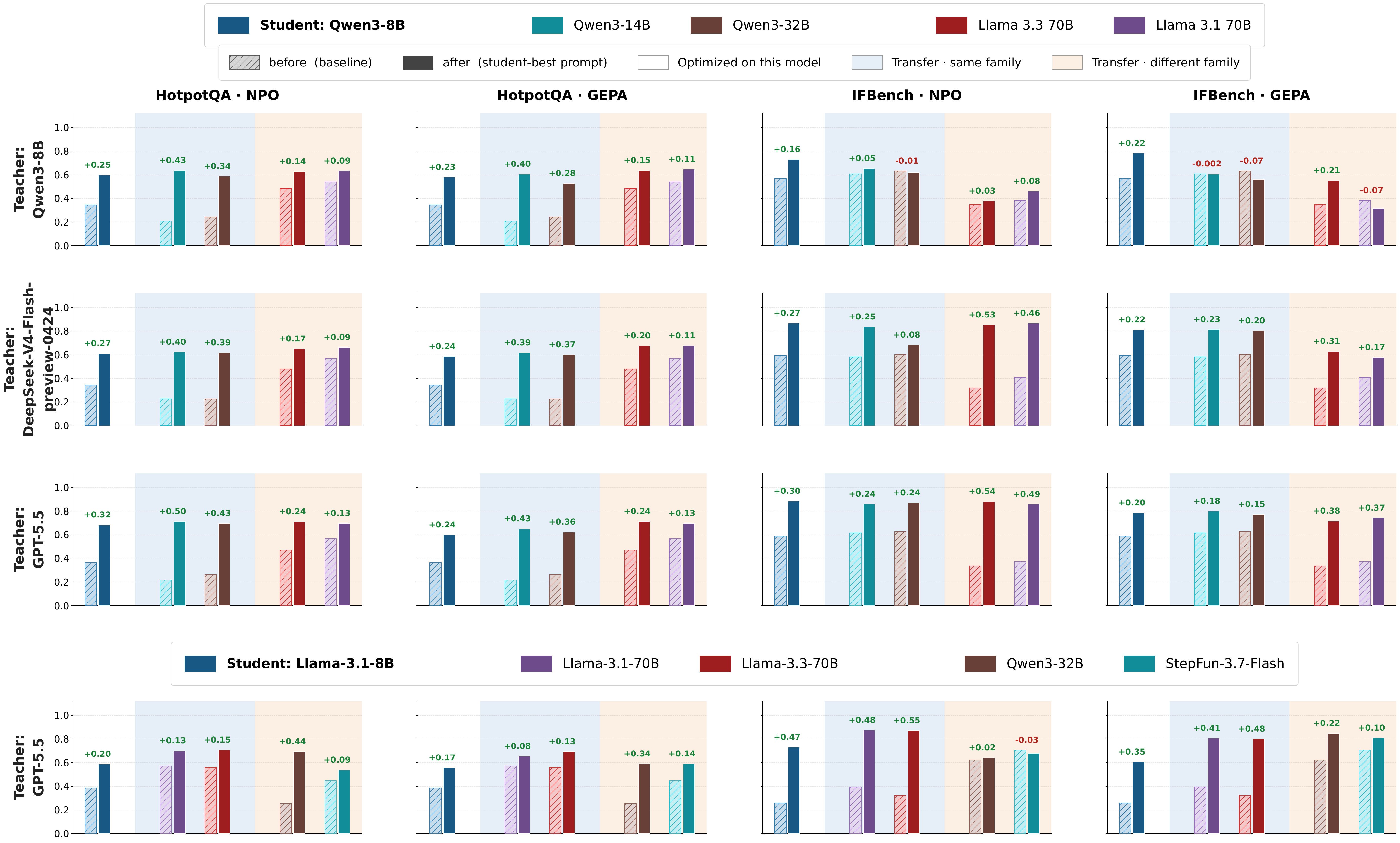}
    \caption{Performance improvements when prompts learned by NPO are transferred verbatim to different student models, within and across model families.
    }
    \label{fig:transfer-across-student}
\end{figure}
Overall, transfer is strongest within model families but remains effective across families.
For both Qwen and Llama, prompts optimized on small models also yield substantial performance gains on larger models within the same family, showing that the benefits of prompt optimization can persist across model scales.
Cross-family transfer also produces meaningful gains, although the improvements are generally slightly weaker and more variable across tasks and optimization methods.

Figure~\ref{fig:transfer-across-student} is organized by tasks and optimization methods across columns and by teachers across rows. 
Each plot compares the performance improvement of the original student model with the performance improvements obtained by applying prompts transferred from the optimization student.
Qwen3-8B and Llama-3.1-8B, which serve as the target students during prompt optimization, are shown in dark blue within the unshaded white region. The light-blue shaded region denotes within-family transfer experiments, whereas the yellow shaded region denotes cross-family transfer experiments. Most settings exhibit positive performance gains, with only a few exceptions, suggesting that transfer performance can still vary across specific model, task, and optimization-method combinations. Detailed optimization trajectories for each teacher--method--dataset combination, including transfer performance across evaluated student models, are provided in Appendix~\ref{app:transfer-trajectories}.

\subsection{Comparison against GRPO}

\begin{figure}[t]
    \centering
    \includegraphics[width=0.95\textwidth]{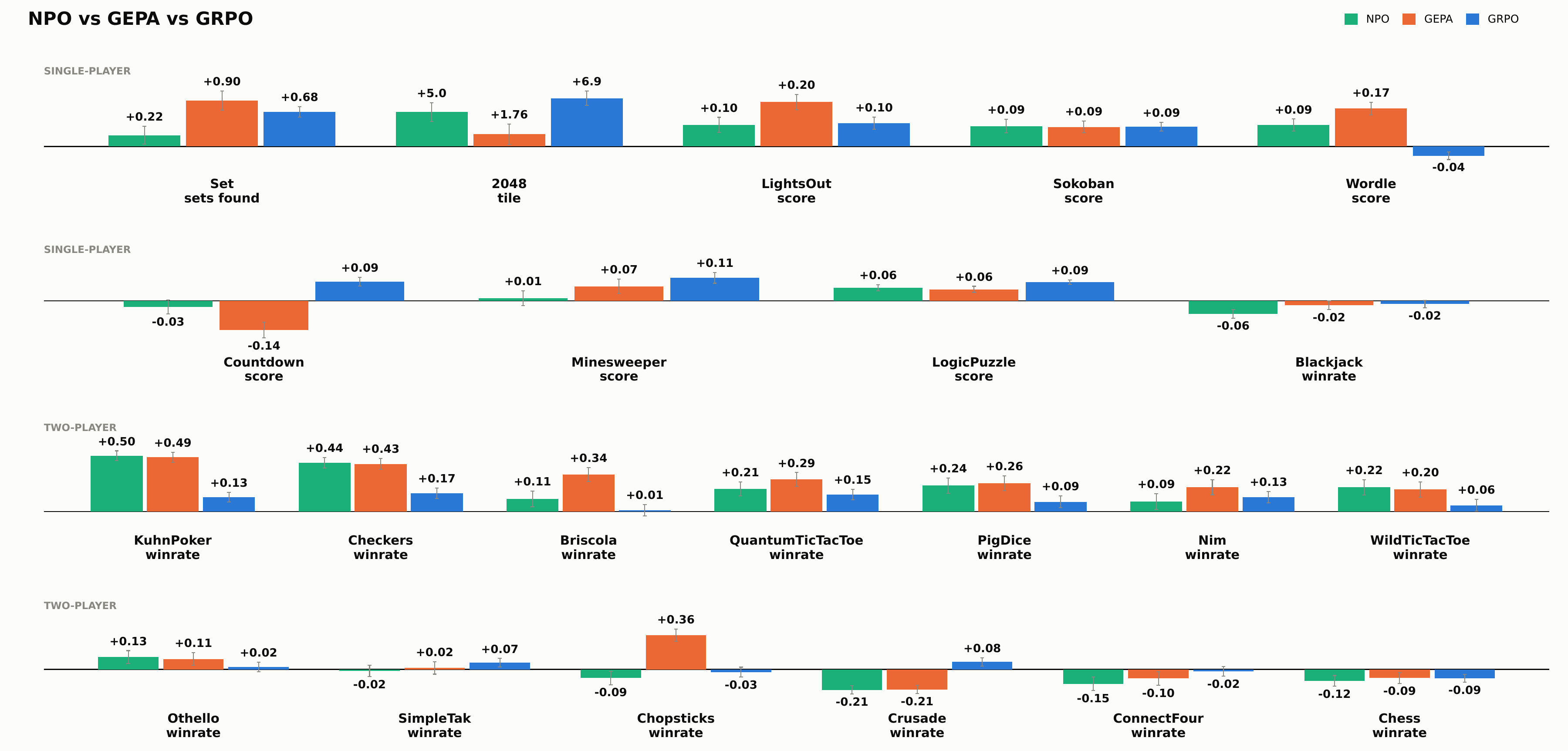}
    \caption{Performance improvement under different optimization methods for 22 TextArena games.}
    \label{fig:barplot-pergame}
\end{figure}

Figure~\ref{fig:barplot-pergame} compares the performance gains of NPO, GEPA, and GRPO across 22 different game environments from TextArena, controlled for the same rollout budget.
The student model is Qwen3-8B, while GPT-5.5 is used as teacher for NPO and GEPA.
The results show no universal winner across the diverse interactive environments.
Contrary to earlier reports, GEPA does not consistently outperform GRPO across all tasks.
NPO achieves similar gains to GRPO on several games, with GEPA showing no consistent advantage over the much simpler NPO method.
Detailed results across all games, evaluation metrics, and optimization methods are provided as heatmaps and per-game plots in Appendices~\ref{app:textarena-heatmap} and~\ref{app:additional-game-plots}.
\subsection{Measuring answer leakage in optimized prompts}

\begin{figure}[t]
    \centering
    \includegraphics[width=\textwidth]{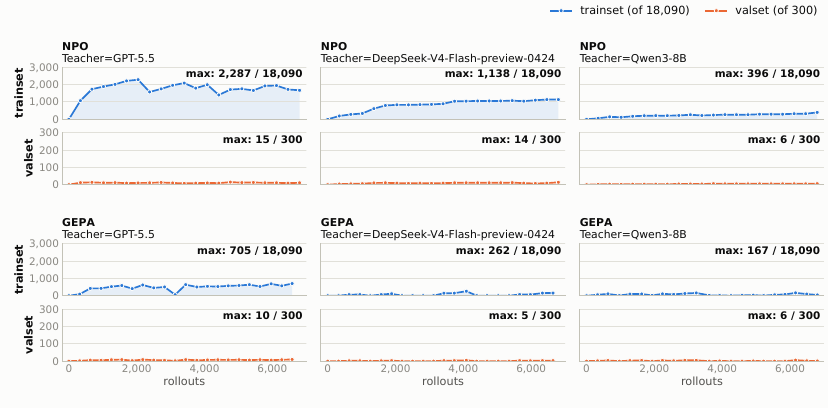}
    \vspace{-1em}
    \caption{Measuring gold-answer leakage into optimized prompts (HotpotQA).
    }
    \label{fig:leakage}
\end{figure}

NPO, particularly when paired with stronger teachers such as GPT-5.5, sometimes produces substantially longer prompts than GEPA. Because the teacher revises prompts using feedback from training rollouts, optimized prompts may naturally incorporate information specific to the training examples. Such incorporation is not necessarily problematic: like few-shot prompting, prompt optimization is intended to extract useful information from training examples and encode it in the resulting prompt. 
However, if gold answers from the held-out validation set had leaked into the teacher model’s training data, become encoded in the teacher’s parametric knowledge, and subsequently been incorporated into an optimized prompt, the measured gains could reflect such contamination rather than improved task-solving ability.

To assess this possibility, we compare every prompt version generated during optimization against gold answers from both the training and validation sets, allowing us to examine whether answer overlap accumulates over successive revisions. As shown in Figure~\ref{fig:leakage}, both NPO and GEPA yield prompts whose overlap with training set answers grows naturally, whereas overlap with validation answers remains negligible. Manual inspection further shows that the small amount of nonzero validation overlap arises from distinct training and validation questions that share the same answer, rather than from direct exposure to validation examples. These results suggest that the observed performance gains are unlikely to be explained by evaluation-answer leakage.

{\setlength{\intextsep}{2pt plus 1pt}
\begin{figure}[t]
    \centering
    \includegraphics[width=0.95\textwidth]{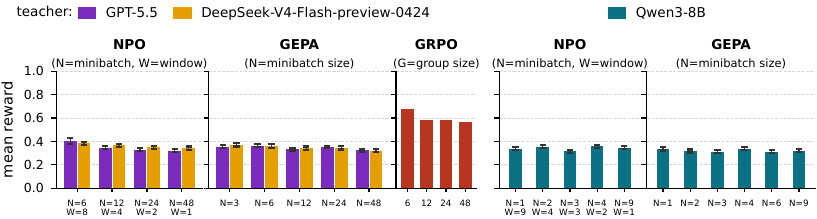}
    \caption{Effect of different reflection minibatch sizes and lookback window sizes (Minesweeper).}
    \label{fig:batch-window-main}
    \vspace{6pt}%
    \centering
    \includegraphics[width=0.55\textwidth]{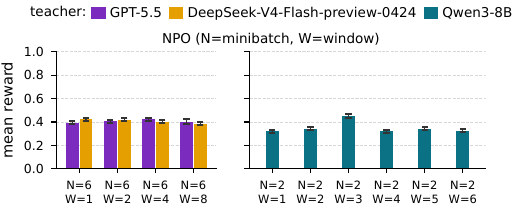}
    \caption{Effect of different lookback window lengths under the same batch size  (Minesweeper).}
    \label{fig:window-size}
    \vspace{6pt}
    \includegraphics[width=0.95\textwidth]{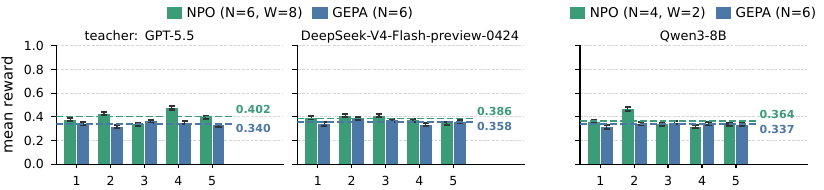}
    \caption{Comparing repeated runs across five different environmental random seeds (Minesweeper).}
    \label{fig:batch-window-seeds}
\end{figure}}

\subsection{Analyzing performance across different minibatch sizes and sliding window sizes}
\label{sec:batch-window}

We now analyze the sensitivity of NPO and GEPA to their key hyperparameter choices: reflection minibatch size and sliding window size. 
Using the Minesweeper game in TextArena, in which we observed NPO and GEPA did not outperform GRPO,
we sweep the hyperparameter space to observe the effect and compare their performance at the same minibatch sizes.
We again use Qwen3-8B as the student and use both stronger models GPT-5.5 and DeepSeek-V4-Flash-preview-0424 and Qwen3-8B itself as teachers, and fix the total rollout budget at 408.
For GEPA we vary the reflection minibatch size between $N\in\{3, 6, 12, 24, 48\}$; for NPO, we use minibatch size and sliding window size combinations  
$(N,W) \in \{(6,8), (12,4), (24,2), (48,1)\}$, so that each reflection contains 48 rollouts, still fitting into the teacher model's 1 million token context window limit. 
For Qwen3-8B as teacher, to accommodate its smaller context window limit that can only fit less than 10 rollouts, we use $(N,W) \in\{(1,9), (2,4), (3,3), (4,2), (9,1)\}$ instead.
As shown in Figure~\ref{fig:batch-window-main}, both NPO and GEPA seem to work best at minibatch size $N=6$ but the difference is minuscule. GEPA performance dropped at the largest reflection minibatch size $N=48$, possibly due to fewer iteration steps given the constant budget of implementation.

For NPO,  we further vary the sliding window size only while keeping a constant minibatch size $N=6$, using $W \in \{1,2,4,8\}$ past batches in each reflection. Figure~\ref{fig:window-size} shows NPO under stronger teachers might benefit slightly from a longer sliding window, although at a higher teacher reasoning cost. Under self-reflection, the result is inconclusive. 

To assess run-to-run variability, we repeat five independent replication runs under different random seeds using GEPA's optimal minibatch size $N=6$ and NPO's optimal sliding window $W=8$ under $N=6$ for strong teachers (and $N=4,W=2$ for self-reflection).
As shown in Figure~\ref{fig:batch-window-seeds}, 
despite having some run-by-run variation, NPO outperforms GEPA on average under different teachers.
We also use GEPA and NPO's best hyperparameters in the context-limited Qwen3-8B as teacher scenario; NPO still shows a better average across runs.
These results suggest that NPO can indeed match or exceed GEPA’s performance.

\subsection{Inference efficiency}
To understand how NPO affects inference cost, we measured the number of prompt and reasoning tokens per call. NPO produced longer prompts than GEPA on average, with 852 (vs. 637) tokens using GPT-5.5 as the teacher, 612 (vs. 473) tokens using DeepSeek-V4-Flash-preview-0424, and 416 (vs. 333) tokens using Qwen3-8B for self-reflection. Meanwhile, NPO elicited slightly shorter reasoning chains than GEPA, averaging 866 (vs. 1,080) tokens with GPT-5.5, 1,056 (vs. 1,356) with DeepSeek-V4-Flash-preview-0424, and 1,333 (vs. 1,463) with Qwen3-8B. As noted earlier in Section~\ref{sec:iteration_settings}, forcing the model to generate more reasoning tokens did not improve reward.

Many newer models use speculative decoding to accelerate token generation~\cite{gloeckle2024mtp}. Prior work reported that speculative decoding token acceptance rates can decline during RL training~\cite{chen2026respec,wang2026aurora,shao2026das}. We therefore also measured how token acceptance rates changed during NPO and GEPA prompt optimization and GRPO training on Qwen3.5-9B, a student model with a multi-token prediction (MTP) module. Mean acceptance rates shifted when prompts were revised, but we observed no sustained increase or decrease during prompt optimization or GRPO training. See Appendix~\ref{app:Speculative-Decoding-MTP} for details.


\section{Discussion and conclusion}

We introduce NPO as a simple iterative baseline to show sophisticated prompt search might not be necessary. On IFBench and HotpotQA, NPO achieves better performance than GEPA with stronger teachers, suggesting that richer trajectory-level feedback can partially compensate for reduced search complexity. Additional experiments varying reflection minibatch size, history window, and per-run variance further support NPO's effectiveness as a lightweight and simple alternative to sophisticated prompt search mechanisms. Optimized prompts also transfer to larger and cross-family student models without re-optimization, although effectiveness varies by model, task, and optimization method.
Across all 22 TextArena games, NPO and GEPA remain broadly competitive, while GRPO provides complementary improvements on tasks less amenable to prompt optimization.

\paragraph{Responsible-Use Statement.}
Our results suggest a strong teacher model might be able to improve student performance for arbitrary tasks with very few rollouts, lowering the cost of rapidly optimizing  small, locally deployed models for malicious tasks; this efficiency could be misused to quickly develop or strengthen malicious agents. A key mitigation is to preserve the teacher model's safety guardrails: the teacher should not only refuse to perform harmful tasks itself, but also avoid generating optimized prompts that would help another model perform them.
Because modern multi-layered AI safeguards commonly include dedicated content- and intent-filtering components, we expect them to exhibit similar refusal behaviors  whether the teacher is asked to perform a harmful task directly or to optimize another model for that task; nevertheless, this behavior should be explicitly verified and monitored in deployment.
NPO also requires sharing the student's rollout traces with the teacher model, which may be hosted on untrusted third-party infrastructure, particularly when a closed model is used as the teacher.
Thus, masking and sanitization are required for tasks involving sensitive data.

\paragraph{Limitations.}
Due to budget constraints, we only evaluated on limited runs on a small set of tasks, leaving open whether NPO's advantage extends to more complex, long-horizon agent environments. We also do not evaluate frontier-scale closed models like GPT-5.5 as students, primarily due to temporal performance inconsistencies and the lack of token-level logits and sufficient fine-grained decoding control exposed via APIs.
In addition, we observe that RL training remains unstable on some tasks and that different tasks may favor different RL algorithms or reward designs. Long-horizon environments further increase both RL instability and the context required to provide NPO teachers with sufficiently large sliding windows over rollout histories; therefore, our results do not yet establish the relative effectiveness of prompt optimization and RL in longer-horizon settings.

\paragraph{Future Directions.}
Building on these preliminary findings, we identify two potential directions for future work. First, for prompt tuning techniques in general, it may be worthwhile to investigate whether prompts optimized independently for multiple diverse tasks can be consolidated into a compact multi-task prompt while preserving their task-specific gains, followed by a few additional rounds of NPO on the consolidated multi-task prompt. 
Second, we propose systematically searching, using larger teacher and smaller student models, for a compact set of representative \textit{eigen-}tasks such that prompts optimized on this small basis can be distilled into a general-purpose prompt that yields substantial improvements across the broader task space, while remaining transferable to larger student models. For example, optimizing prompts on a basis consisting of a particular board game, a card game, and a knowledge-retrieval task might yield a prompt that produces generalizable improvements across a diverse set of many different board games, card games, and knowledge-retrieval tasks.

\paragraph{Acknowledgments.}
We are grateful to Jason Cheuk Nam Liang for his helpful discussions and feedback. We thank Chengming Liao for providing OpenAI API credits that supported this work.

\bibliographystyle{plainnat}
\bibliography{references}

\clearpage
\appendix

\raggedbottom

\newcommand{\appendixfigure}[5][0.94\textwidth]{%
    \par\noindent
    \begin{minipage}{\textwidth}
        \centering
        \includegraphics[
            width=#1,
            height=#3,
            keepaspectratio
        ]{#2}
        \par\vspace{3pt}
        \refstepcounter{figure}\label{fig:#5}
        \parbox{#1}{%
            \centering\small
            Fig.~\thefigure. #4
        }
    \end{minipage}
    \par
}

\section{Detailed optimization trajectories in cross-student transfer experiments}
\label{app:transfer-trajectories}

\begin{figure}[H]
    \centering
    \includegraphics[width=\textwidth]{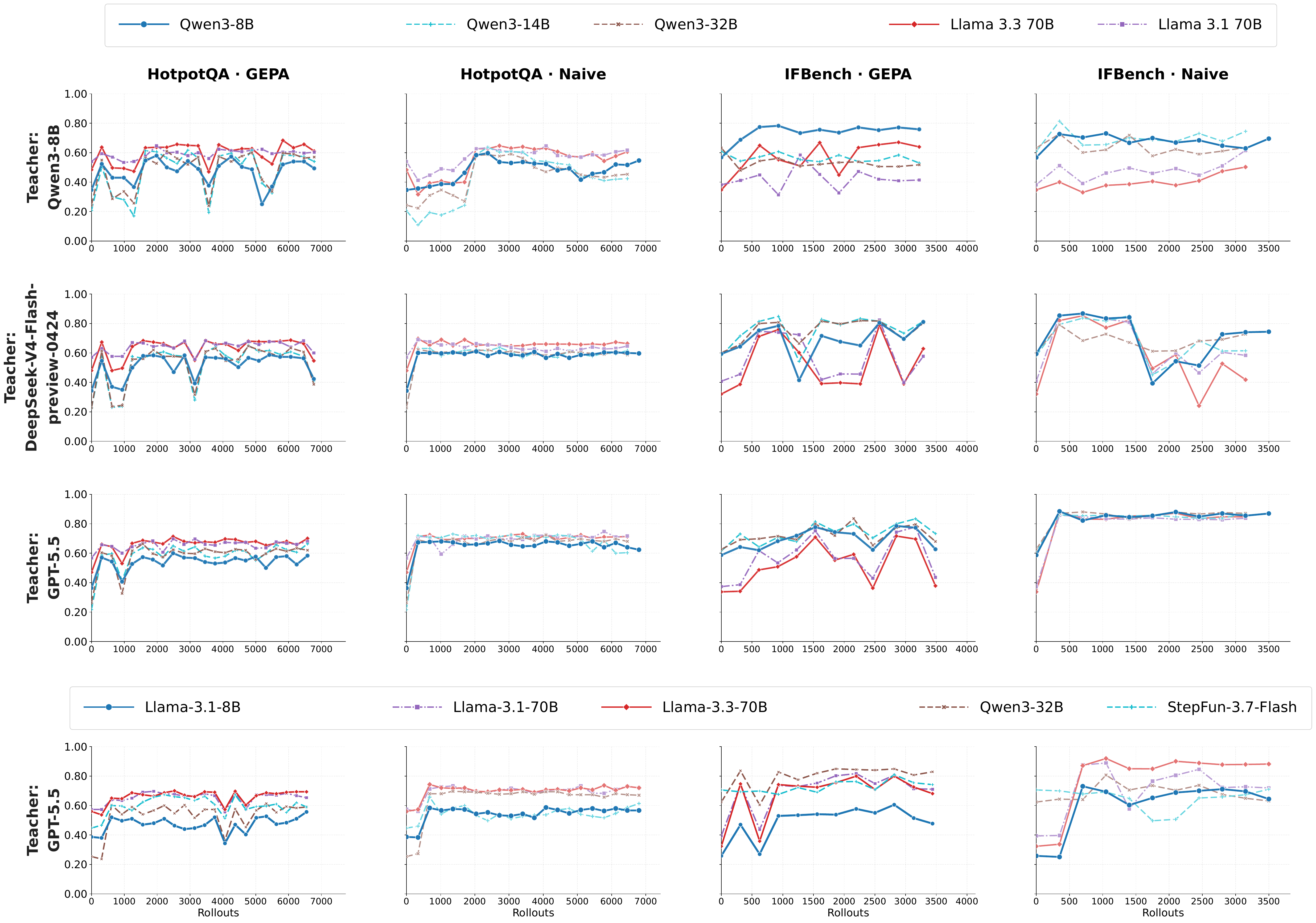}
    \caption{Optimization trajectories for the cross-student transfer experiments. Columns correspond to dataset--optimization-method combinations; rows correspond to teacher models.}
    \label{fig:trajectories}
\end{figure}

Figure~\ref{fig:trajectories} presents the detailed optimization trajectories for the cross-student transfer experiments in IFBench and HotpotQA. Columns correspond to different dataset--optimization-method combinations, while rows correspond to different teacher models. Each curve shows how prompts from successive NPO iterations or candidates from the GEPA pool transfer to larger and cross-family student models, leading to similar performance gains.

\vspace{8pt}

\vspace{2pt}

\vspace{15pt}
\section{Heatmap for the evaluated games in TextArena}
\label{app:textarena-heatmap}
Figure~\ref{fig:heatmap} contains six sections: Sections~1--3 show the unoptimized Qwen3-8B baseline, GRPO results, and held-out NPO/GEPA performance using DeepSeek-V4-Flash-preview-0424 and GPT-5.5 as teachers. GRPO achieves strong gains on several tasks with larger budgets and carefully designed rewards, while NPO and GEPA remain broadly comparable. Sections~4--6 evaluate transfer by applying prompts optimized on Qwen3-8B unchanged to Llama-3.3-70B-Instruct, Qwen3-14B, and Qwen3-32B. Same-family Qwen transfer largely preserves performance gains, while cross-family transfer to Llama-3.3-70B-Instruct is less consistent. Cells report average performance for each game--method combination; win rates are shown as decimals.

\begin{center}
\makebox[\textwidth][c]{%
    \includegraphics[
        width=1.02\textwidth
    ]{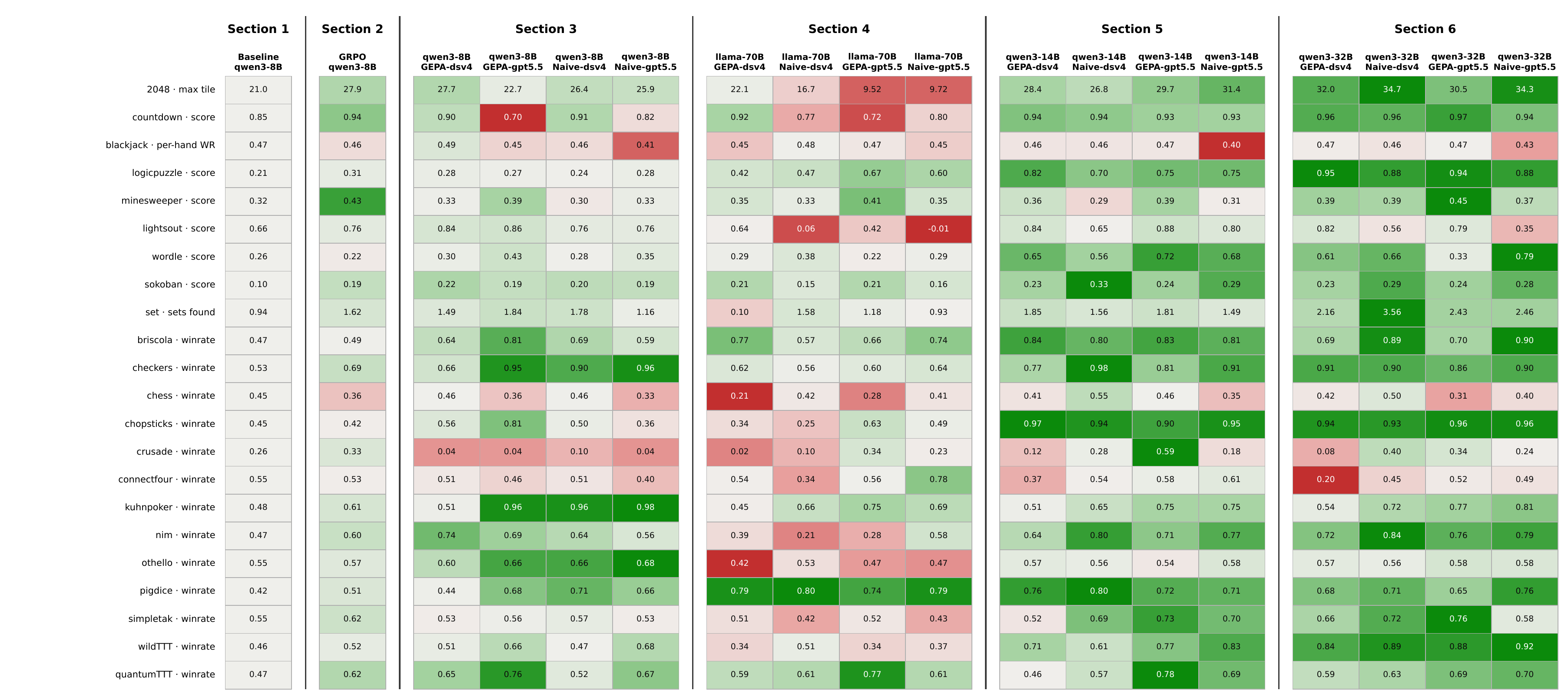}%
}

\vspace{3pt}

\refstepcounter{figure}\label{fig:heatmap}
{\small Fig.~\thefigure. Performance heatmap across 22 TextArena games for prompt optimization, weight-based RL fine-tuning, and cross-student prompt transfer.}
\end{center}

\vspace{8pt}

\section{Extended analysis of GRPO group size}
\label{app:grpo-group-size}

We conducted a small experiment to examine how rollout group size affects GRPO training quality on TextArena games. We first supervised-fine-tuned Qwen3-8B on GPT-4o-mini responses~\citep{instructgpt,lora}, which typically contain about 300 tokens. 
We then trained the SFT-initialized model with GRPO under a fixed setup using different rollout group sizes. 
We show the results in Minesweeper, with other environments leading to similar results.
As shown below in Figure~\ref{fig:groupsize}, larger group sizes did not lead to significantly better training, and the model's skill improvement behaved similarly under the same rollout budget. This might be because the significant variation in TextArena's reward signal means that a very large group size is not required for group-relative normalization to take effect.
This observation is consistent with prior work reporting similar group-size effects~\citep{aero}, as well as broader evidence that larger batch sizes might not always be beneficial in RL training~\citep{smallbatchrl}.

\begin{center}
\makebox[\textwidth][c]{%
    \includegraphics[
        width=1.0\textwidth
    ]{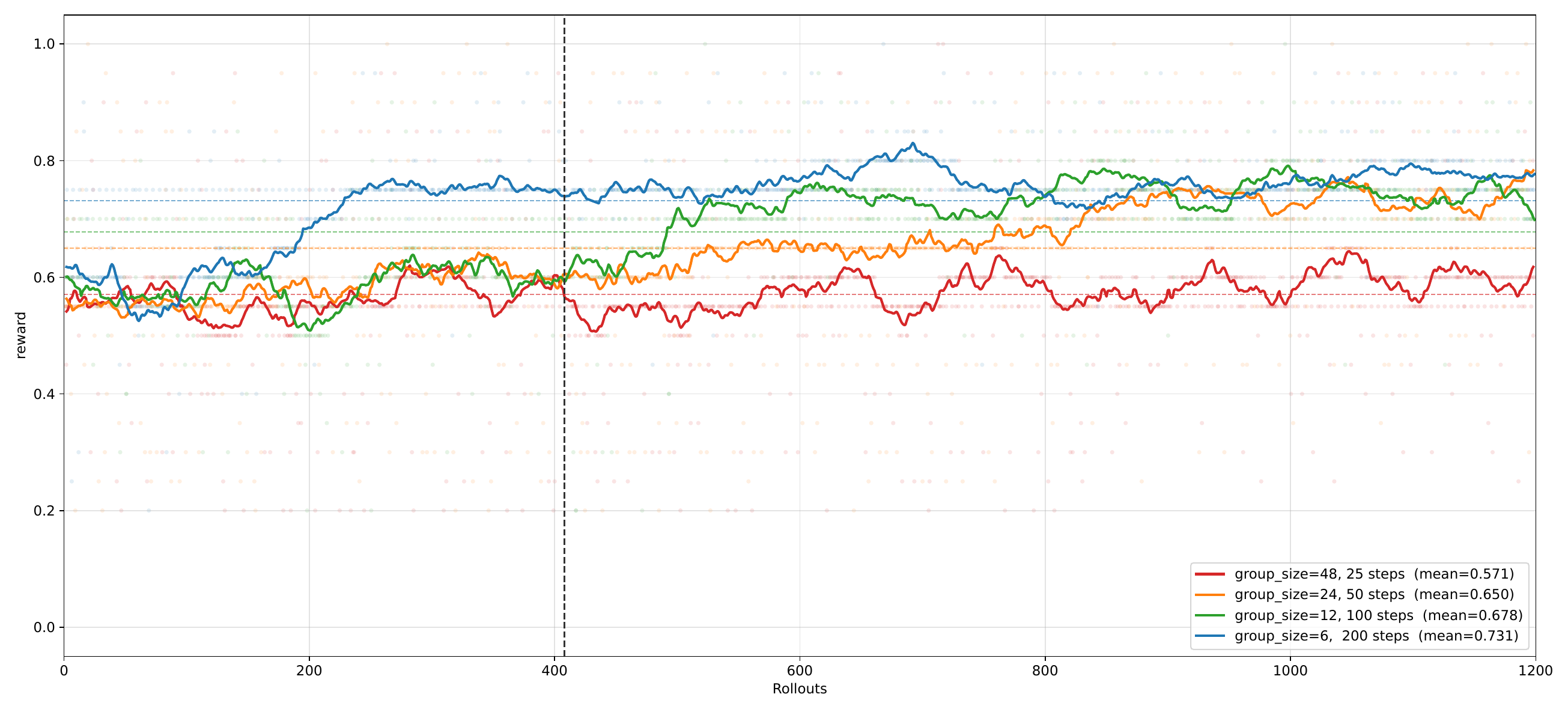}%
}

\vspace{3pt}

\refstepcounter{figure}\label{fig:groupsize}
{\small Fig.~\thefigure. GRPO training performance under different rollout group sizes.}
\end{center}

\section{Constrained decoding implementation}
\label{app:constrained-decoding-design}

\begin{figure}[!htbp]
    \centering
    \includegraphics[width=0.8\linewidth]{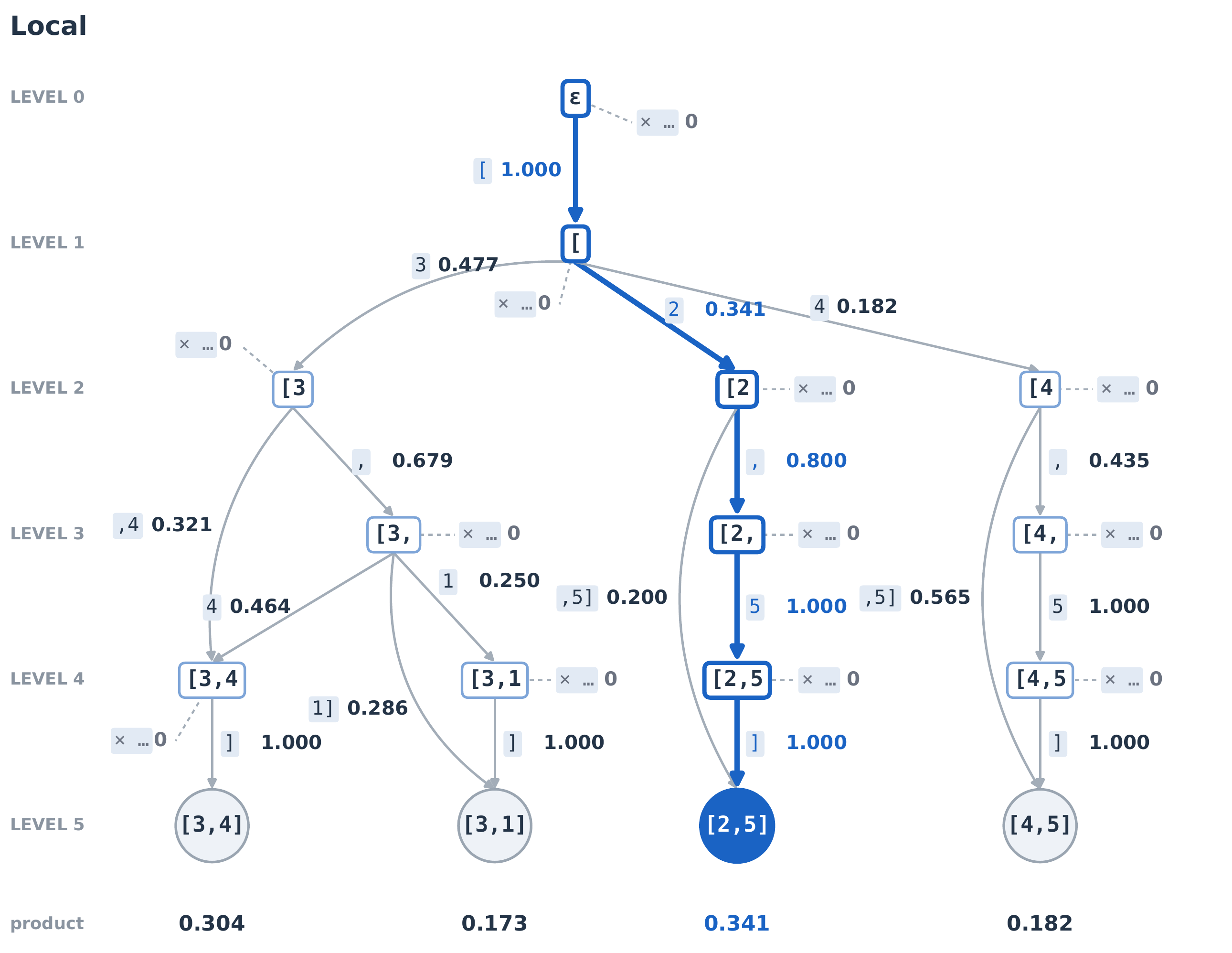}
    \caption{
    Token-prefix trie used for constrained decoding over the set of legitimate action strings. At each node, next-token probabilities are renormalized over the valid outgoing edges that lead to legitimate actions.}
    \label{fig:constrained-decoding-tree}
\end{figure}

We describe how we use constrained decoding to sample an action from the set of legitimate actions supplied by the environment.
Given a finite set of legitimate action strings, we organize all token sequences that decode to these strings into a prefix trie, in which each node represents a string prefix, each edge represents a token, and each leaf represents a complete legitimate action string, as illustrated in Figure~\ref{fig:constrained-decoding-tree}.

At each node with prefix $s$, we mask tokens that are not legitimate continuations and locally renormalize the logits over the remaining outgoing edges (legitimate next-token set) $A(s)$, following standard token-level constrained decoding~\citep{loula2025syntactic}:
\[
q(t\mid s)=\frac{\exp z_t(s)}{\sum_{u\in A(s)}\exp z_u(s)},\qquad t\in A(s).
\]
The probability of a leaf is the product of the locally normalized probabilities along its root-to-leaf path, with probabilities summed across paths if multiple token sequences decode to the same action string.

There are two approaches to computing probabilities over the leaf nodes. In \emph{locally constrained decoding}, also referred to as a local product of experts, outgoing-edge probabilities are normalized at each node before being multiplied along each root-to-leaf path~\citep{loula2025syntactic}. In a \emph{globally conditioned} distribution, the base-model per-token probabilities along each complete path are multiplied first, followed by a single normalization across all legitimate leaves~\citep{park2024grammaraligned}.

We use the former, locally renormalized formulation throughout our evaluation to prevent mechanically required formatting tokens from influencing the relative probabilities assigned to actions.
TextArena actions often contain formatting tokens required by the environment's output syntax.
For example, an action such as \texttt{[3,4]} contains delimiters including \mbox{\texttt{"["}, \texttt{","}, and \texttt{"]"}}. 
At any prefix where only one such formatting token is permitted, locally constrained decoding assigns the corresponding forced transition probability \(1\), effectively eliminating its effect on choices between actions.
Under global conditioning, by contrast, each leaf retains the base-model probabilities assigned to mandatory formatting tokens like \texttt{"]"}. The probabilities of these mandatory formatting tokens can exhibit spurious variation due to preceding choices in the partial action string, causing the same required delimiter to receive different probabilities after different action prefixes. This variation reflects only the model's prefix-dependent likelihood of format compliance, yet it unnecessarily alters each leaf's path probability under global conditioning.

%
%

%

\newcommand{\gamepanel}[2]{%
    \begin{minipage}[t]{0.495\linewidth}
        \centering

        \includegraphics[
            width=\linewidth,
            height=0.235\textheight,
            keepaspectratio
        ]{plots_view/#1/combined_4.pdf}

        \vspace{-1pt}

        {\small\textbf{#2}}
    \end{minipage}%
}

%

\newcommand{\gamerow}[4]{%
    \noindent
    \gamepanel{#1}{#2}%
    \hfill
    \gamepanel{#3}{#4}%
}


\section{Inference efficiency}
\label{app:Speculative-Decoding-MTP}

Since inference engines today often use Speculative Decoding to accelerate auto-regressive generation through next-token predictions, we need to measure whether the prompt-based or weight-based optimizations cause degradation in inference performance due to their effects on next-token prediction accuracy.

We measured inference efficiency through experiments using  Qwen3.5-9B as the student model, which includes 
a pretrained MTP module, through prompt optimization (NPO and GEPA) and GRPO training.
We run experiments and report aggregated statistics across 5 games (Minesweeper, Othello, Nim, Bandit, SimpleTak).

\begin{figure}[h]
    \centering
    \includegraphics[width=\linewidth]{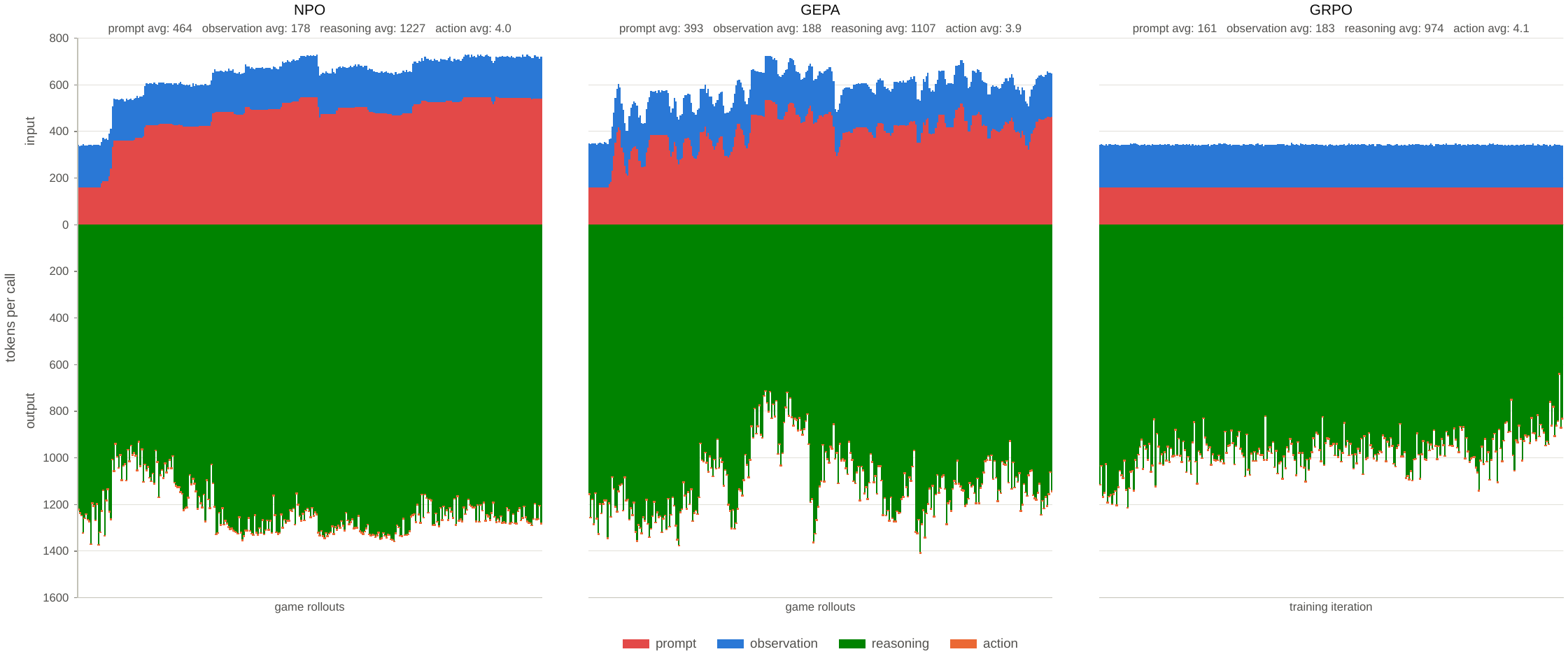}
    \caption{Aggregated token counts for fixed prompts, observations, and generated outputs across the evaluated game tasks in NPO (left), GEPA (middle) and GRPO training (right)}
    \label{fig:mtp_token_counts}
\end{figure}

First, we show the average length of input and output tokens in Figure~\ref{fig:mtp_token_counts}. The input tokens (observation and prompt) are processed faster in prefill; length of environmental observation is the same across
different methodologies, and we notice NPO produces slightly longer prompts than GEPA (averaging 464 and 393 tokens, respectively; 18.1\% longer), similar to earlier experiments where we use Qwen3-8B as the student. GRPO uses the environments’ default prompt (averaging 161
tokens). Output tokens are more costly to generate;  NPO yielded reasoning chains that are only slightly longer than GEPA (averaging 1227 and 1107 tokens respectively; 10.8\% longer).
Since we used constrained decoding, the action token
length is fixed and negligible.

\begin{figure}[h]
    \centering
    \includegraphics[width=\linewidth]{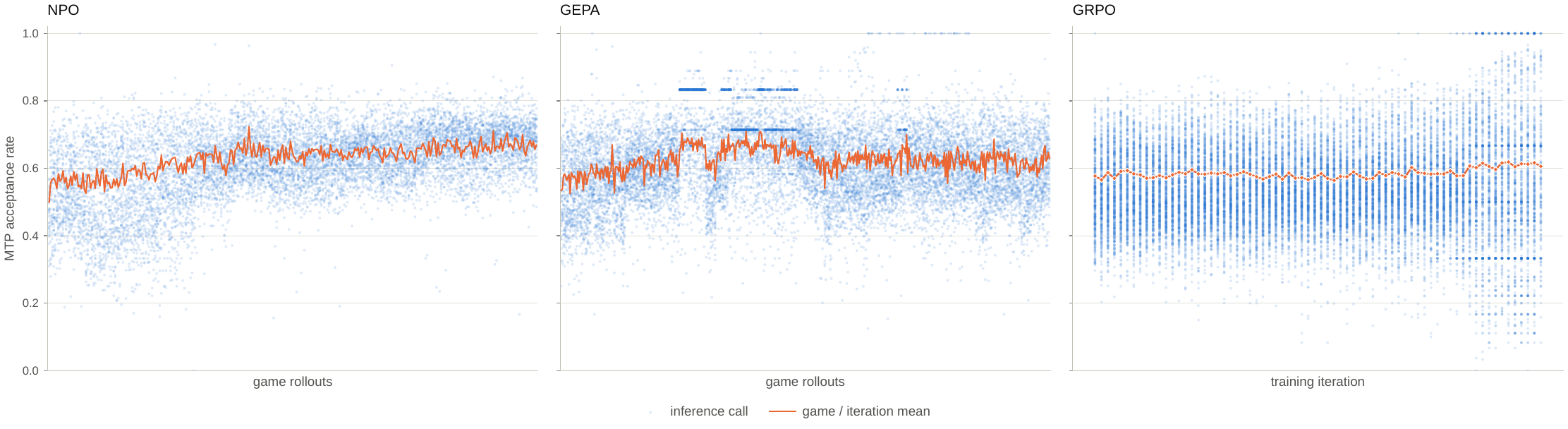}
    \caption{MTP token acceptance-rate trends during training across the evaluated game tasks using Qwen3.5-9B-MTP in NPO (left), GEPA (middle) and GRPO training (right)}
    \label{fig:mtp_acceptance_rate}
\end{figure}

We then measure speculative decoding efficiency via MTP acceptance rates, which directly affect decoding throughput. In our experiment we set MTP to predict the next 3 tokens, and observe the average fraction of tokens accepted.
As shown in Figure~\ref{fig:mtp_acceptance_rate}, our preliminary findings suggest that MTP heads pretrained alongside their base models remain robust to GRPO, with acceptance rates fluctuating but exhibiting no consistent improvement or degradation. 
During prompt optimization, mean MTP acceptance rates varied across prompt versions, with some prompts leading to very high acceptance rates but no consistent trend across iterations or clear association with task rewards.
Overall, we find no statistically significant evidence that either prompt optimization or GRPO systematically improves or degrades MTP token acceptance rates.
We therefore expect similar inference-time throughput when using NPO compared to GEPA and GRPO, given the similar output length and similar MTP acceptance.

\section{Detailed optimization trajectories}
\label{app:additional-game-plots}

This appendix presents the complete optimization trajectories for all 22 TextArena environments. For each game, NPO, GEPA, and GRPO are overlaid on shared axes, with the cumulative number of rollouts on the horizontal axis and the game-specific training metric on the vertical axis. These trajectories provide a direct comparison of the optimization dynamics and relative performance improvements achieved by the three methods across diverse interactive environments.


\vspace{0.6em}

\begin{figure}[H]
    \centering

    \gamerow
        {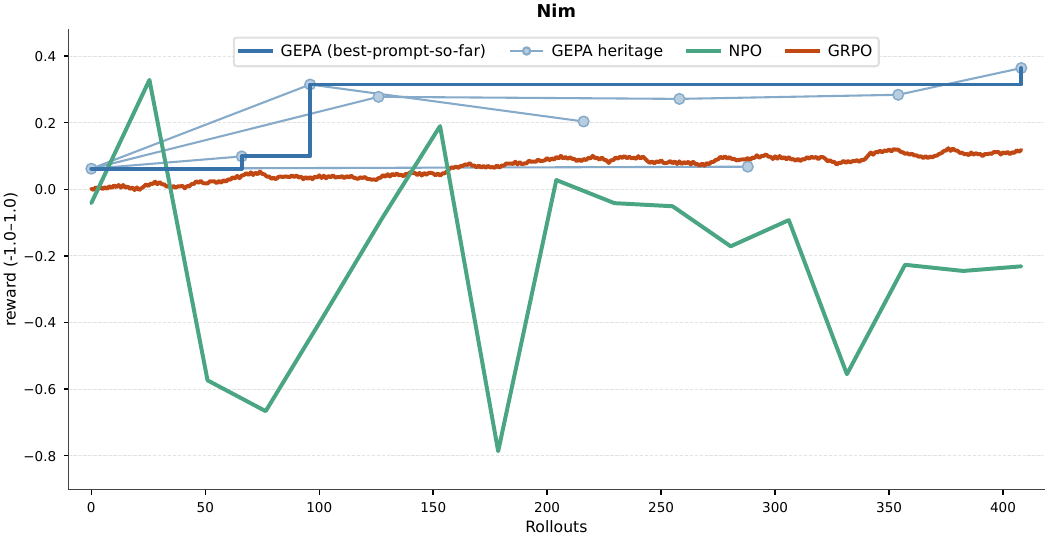}{Nim}
        {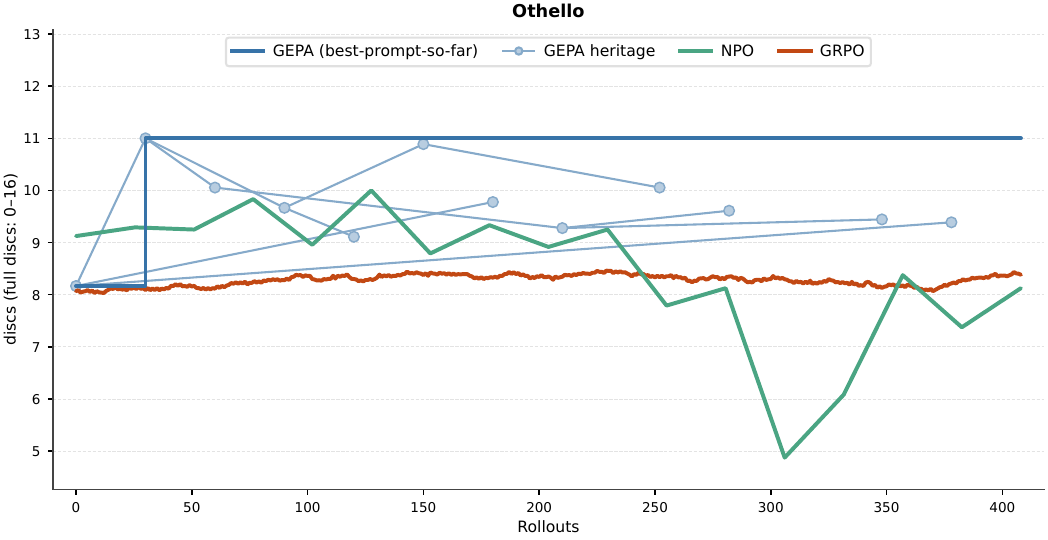}{Othello}

    \vspace{0.8em}

    \gamerow
        {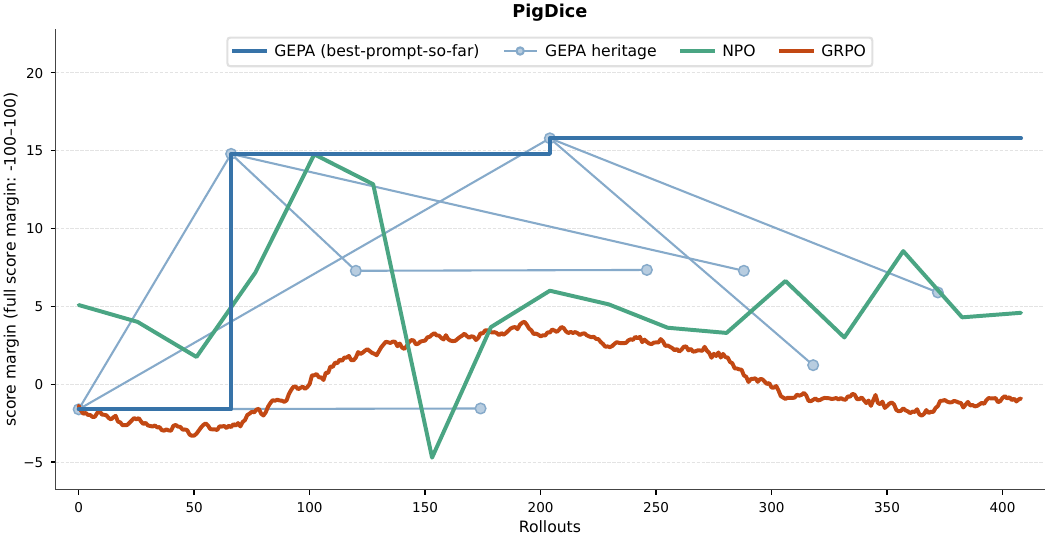}{Pig Dice}
        {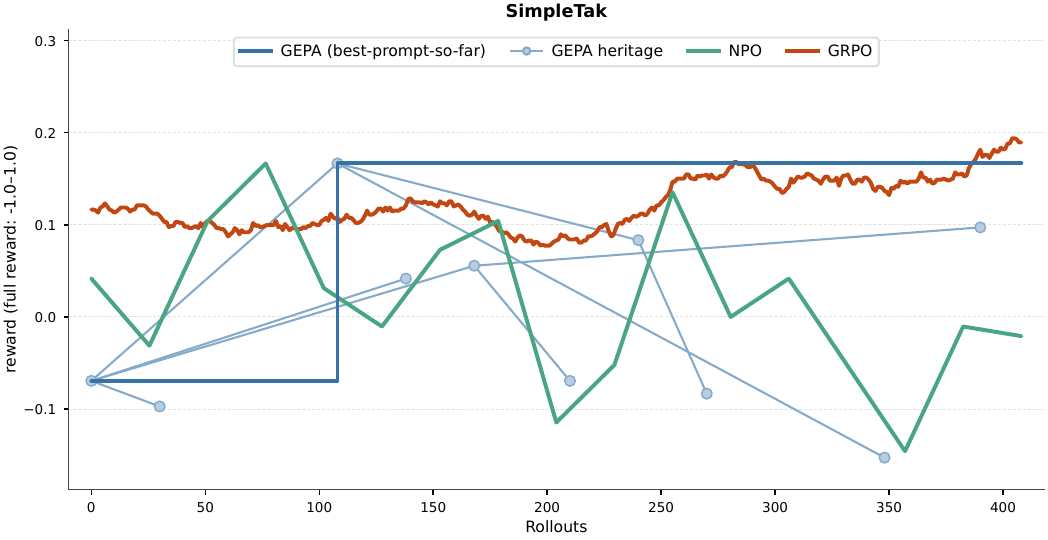}{SimpleTak}

    \vspace{0.8em}

    \gamerow
        {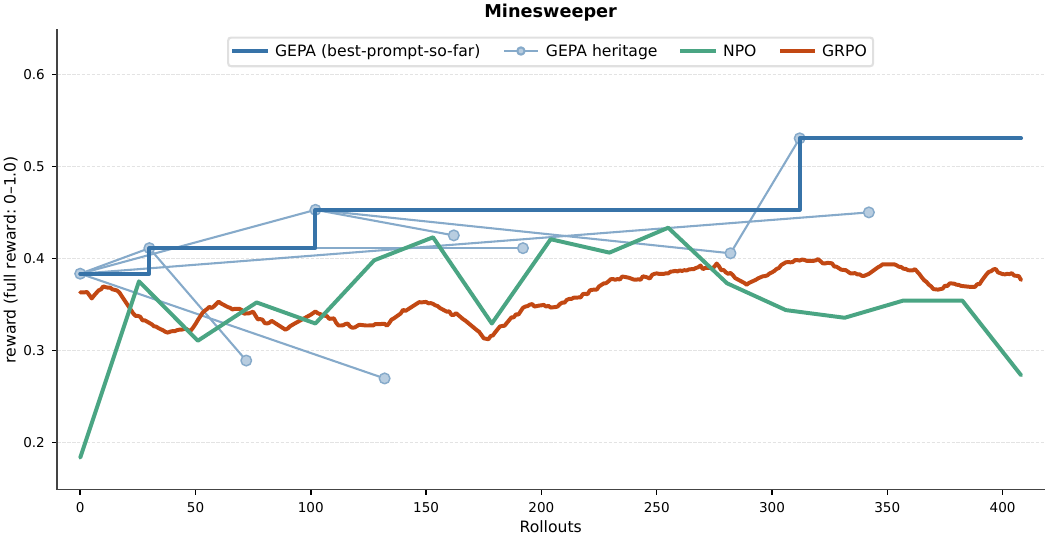}{Minesweeper}
        {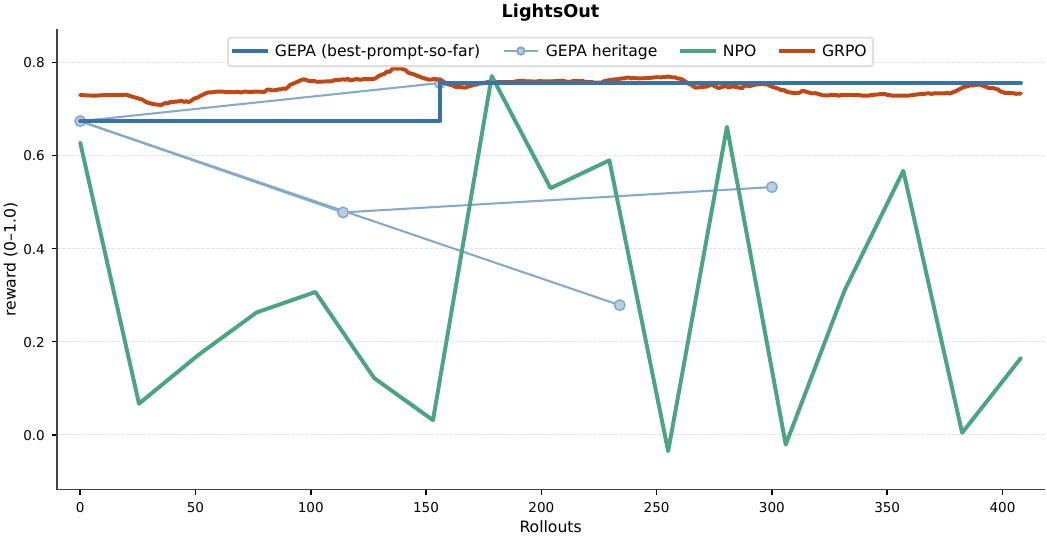}{Lights Out}

    \vspace{0.3em}

    \caption{Performance trajectories of NPO, GEPA, and GRPO across different games.}
    \label{fig:bigfigure}
\end{figure}


\clearpage

\begin{figure}[H]
    \ContinuedFloat
    \centering

    \begin{minipage}[t][0.92\textheight][t]{\linewidth}
        \centering


        \gamerow
            {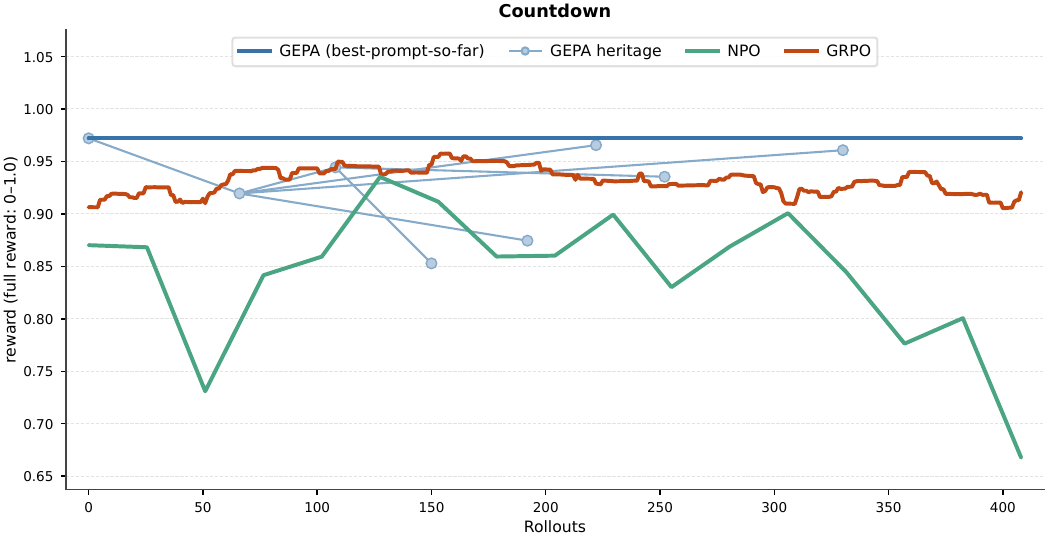}{Countdown}
            {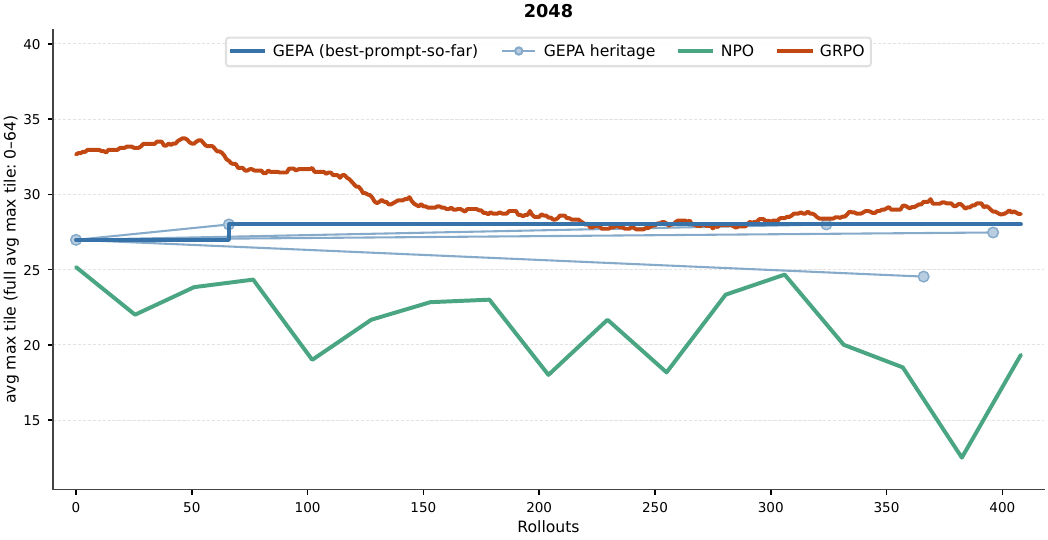}{2048}

        \vfill


        \gamerow
            {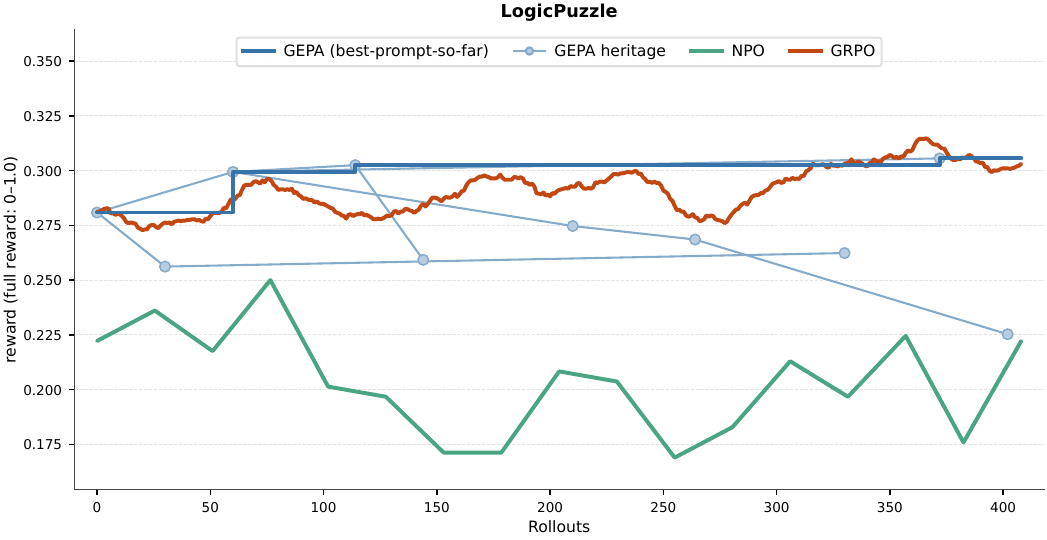}{Logic Puzzle}
            {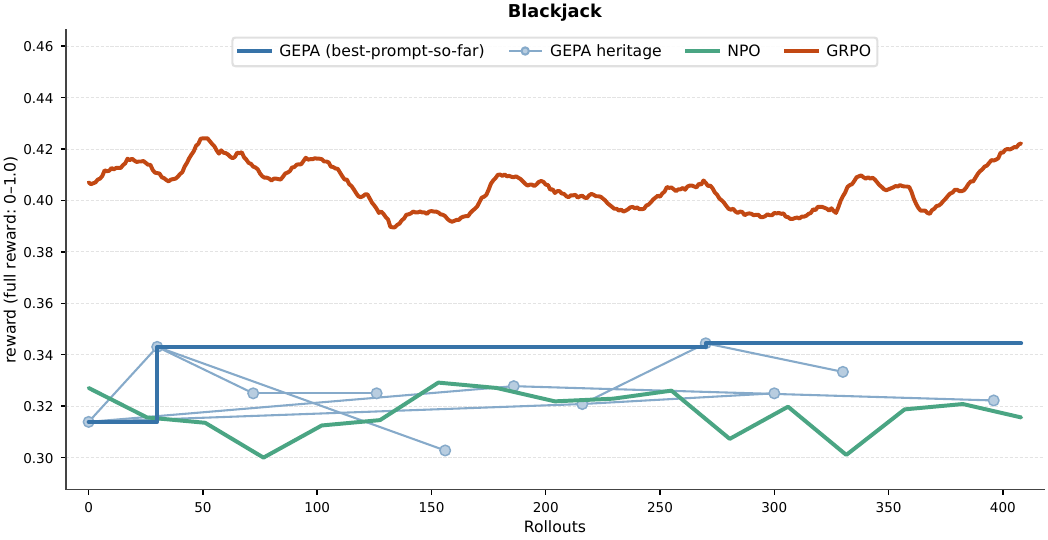}{Blackjack}

        \vfill


        \gamerow
            {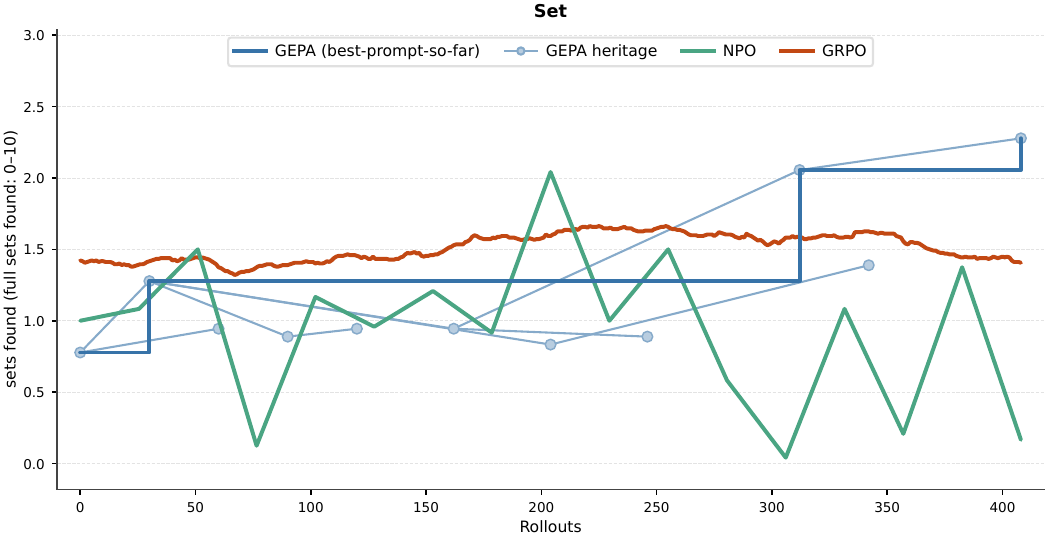}{Set}
            {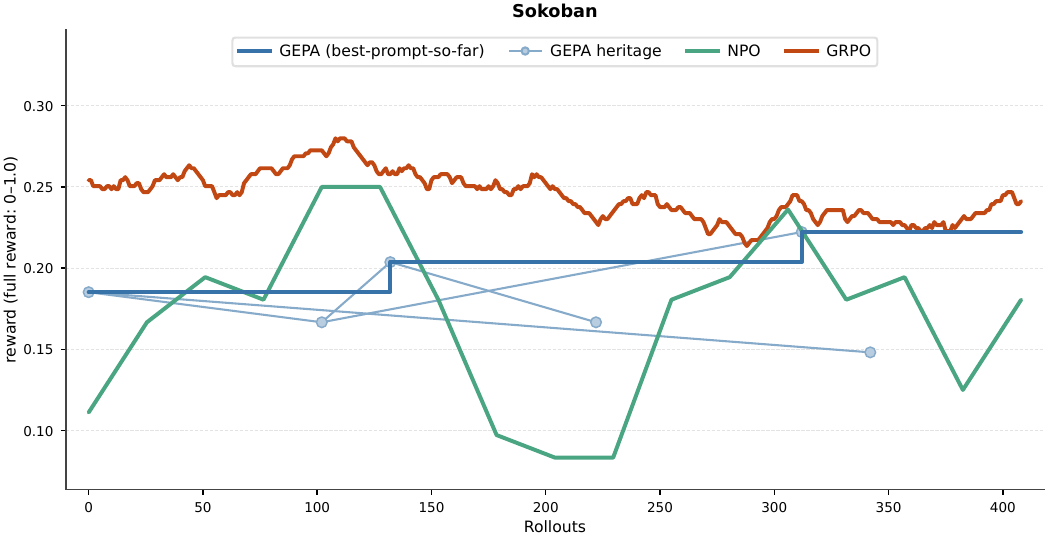}{Sokoban}

        \vfill


        \gamerow
            {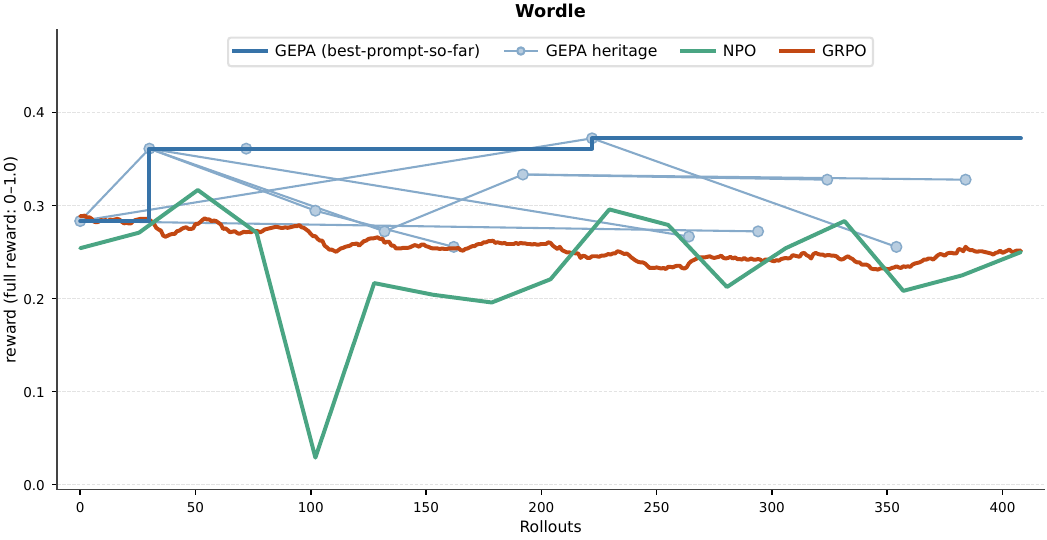}{Wordle}
            {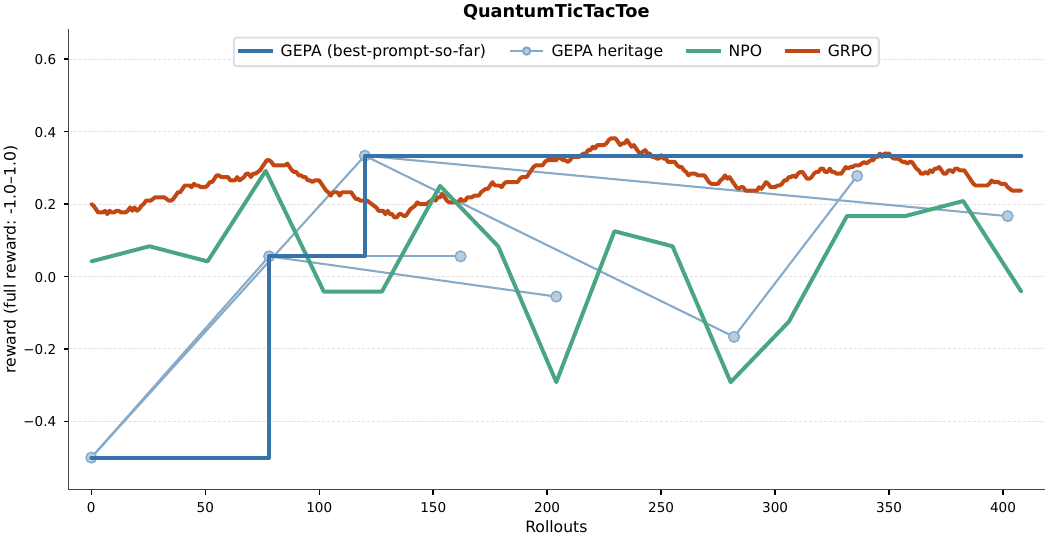}{Quantum Tic-Tac-Toe}

    \end{minipage}

    \caption{Performance trajectories of NPO, GEPA, and GRPO across different games (continued).}
\end{figure}


\clearpage

\begin{figure}[H]
    \ContinuedFloat
    \centering

    \begin{minipage}[t][0.92\textheight][t]{\linewidth}
        \centering


        \gamerow
            {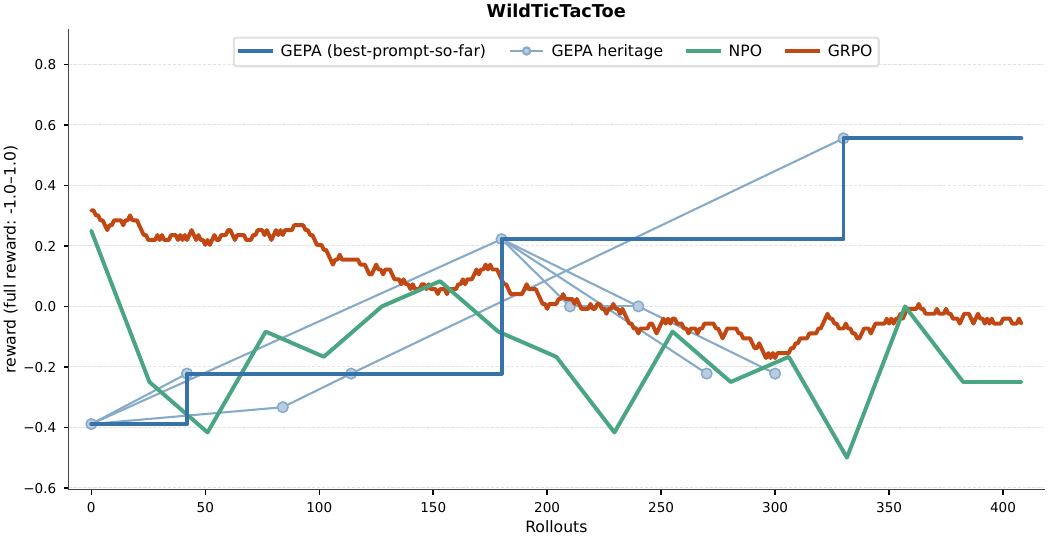}{Wild Tic-Tac-Toe}
            {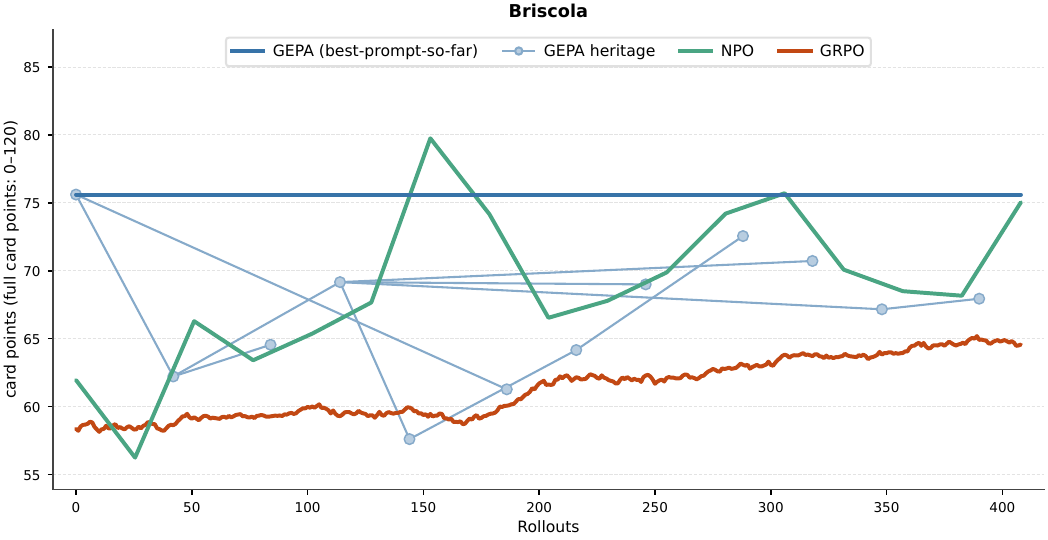}{Briscola}

        \vfill


        \gamerow
            {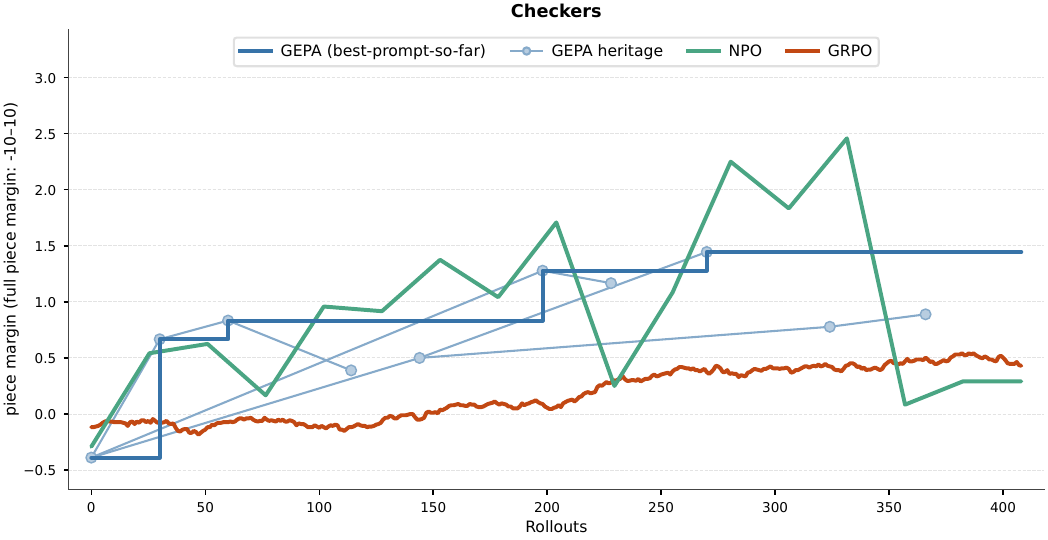}{Checkers}
            {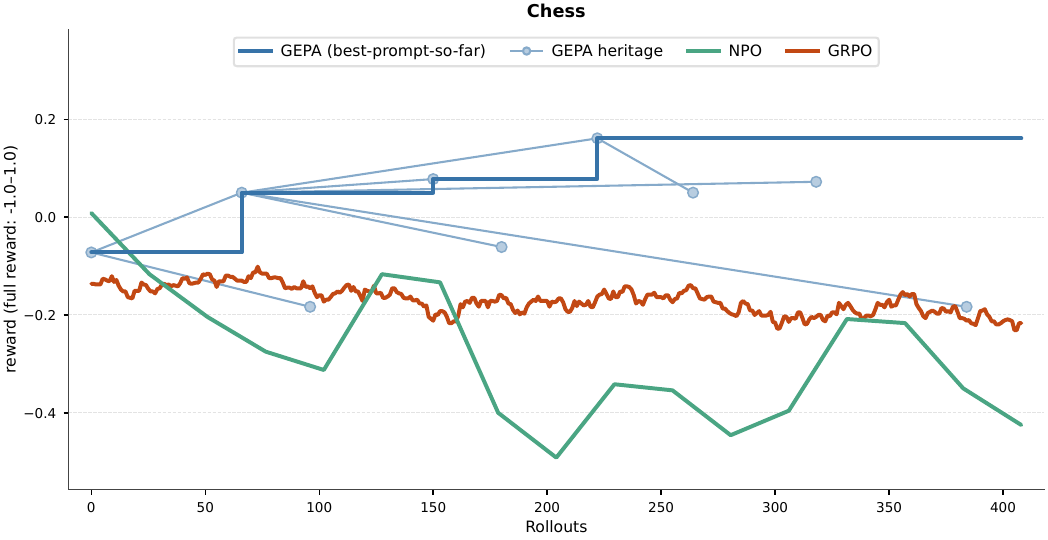}{Chess}

        \vfill


        \gamerow
            {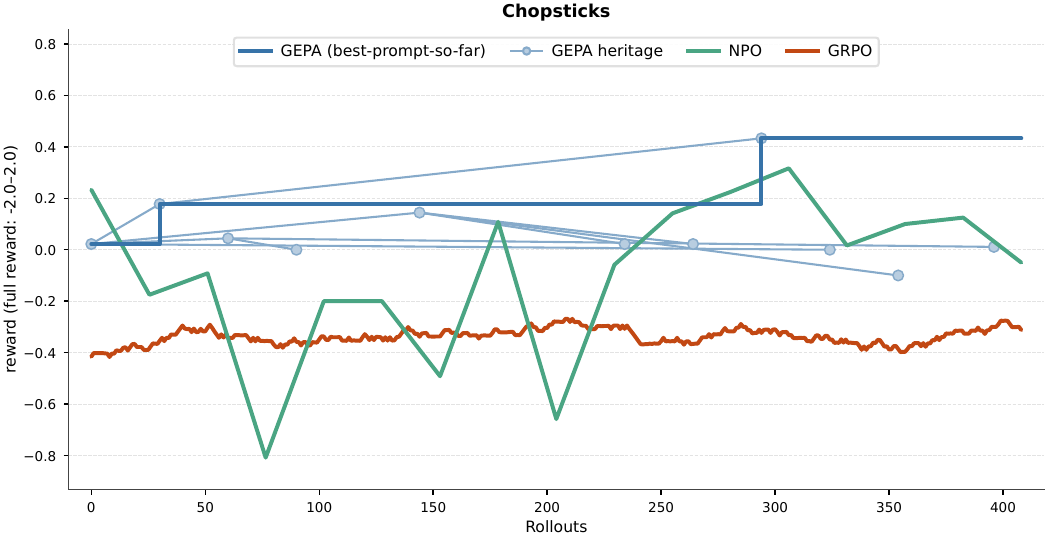}{Chopsticks}
            {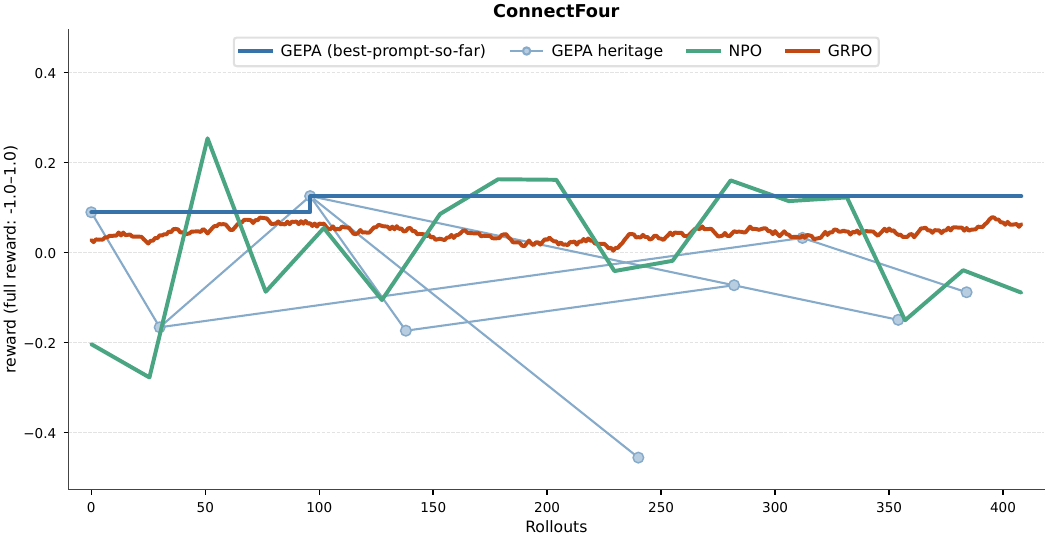}{Connect Four}

        \vfill


        \gamerow
            {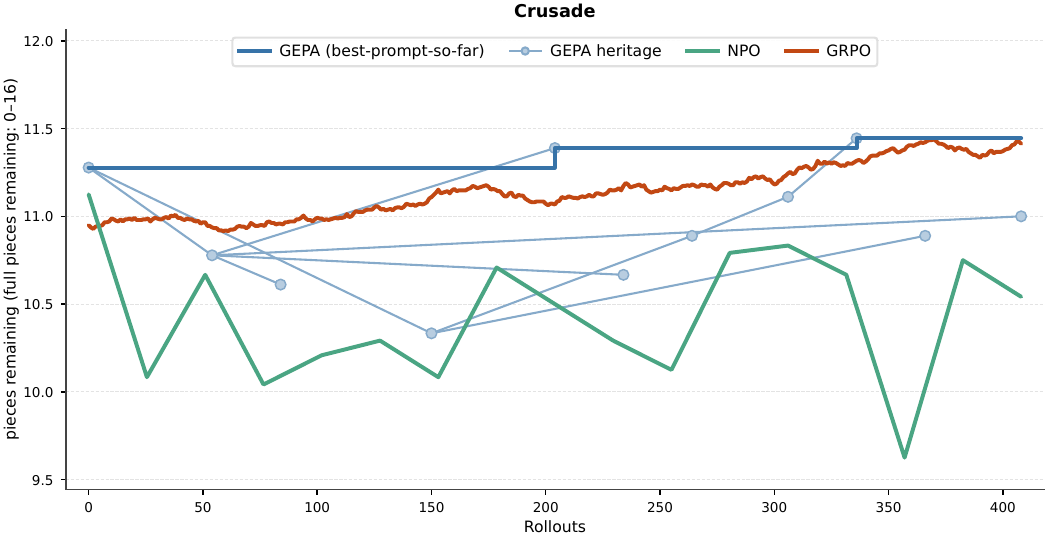}{Crusade}
            {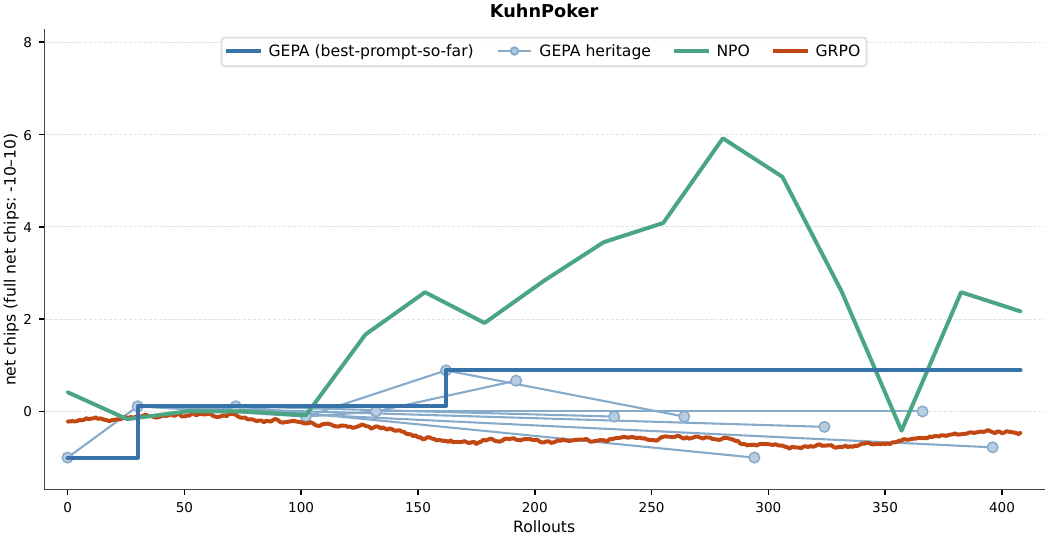}{Kuhn Poker}

    \end{minipage}

    \caption{Performance trajectories of NPO, GEPA, and GRPO across different games (continued).}
\end{figure}

\clearpage

\section{Optimized prompts at a glance}
\label{app:prompts}

In Table~\ref{tab:opt-prompt-excerpts}
we report the optimized prompts for each task produced using GPT-5.5 as the teacher and Qwen3-8B as the student, under GEPA and NPO, respectively. 
Note that HotpotQA and IFBench are multi-hop question-answering and instruction-following tasks, respectively, and the prompts for all stages of both pipelines are optimized jointly.

Despite using fewer rollouts, NPO appears to produce prompts similar in detail and depth to those produced by GEPA.

\vspace{3em}

\begingroup
\setlength{\LTpre}{4pt}\setlength{\LTpost}{4pt}
\setlength{\LTcapwidth}{\dimexpr\textwidth-2em\relax}
\renewcommand{\arraystretch}{1.0}
\begin{longtable}{>{\raggedright\arraybackslash}p{0.155\textwidth}%
                 >{\raggedright\arraybackslash}p{0.038\textwidth}%
                 >{\scriptsize\raggedright\arraybackslash}p{0.715\textwidth}}
\caption{Optimized prompts for each task produced by NPO and GEPA.}\label{tab:opt-prompt-excerpts}\\
\toprule
\multicolumn{2}{l}{\small\bfseries Task / Game} & {\small\bfseries Optimized prompt} \\
\midrule
\endfirsthead
\caption[]{Optimized prompts for each task produced by NPO and GEPA (continued).}\\
\toprule
\multicolumn{2}{l}{\small\bfseries Task / Game} & {\small\bfseries Optimized prompt} \\
\midrule
\endhead
\bottomrule
\endlastfoot

{}\textbf{Hotpot\allowbreak{}QA}\newline \textcolor{GEPAname}{\bfseries GEPA}
& {\scriptsize hop1} &
{}Given the fields \textasciigrave{}question\textasciigrave{} and \textasciigrave{}passages\textasciigrave{}, produce the field \textasciigrave{}summary\textasciigrave{}. Your task is to answer the question accurately and concisely using the provided passages and, when necessary, relevant factual knowledge implied by the question. This is typically a multi-hop question-answering task: identify the entity being asked about, connect it through~{\itshape\textcolor{SelGray}{\ldots(truncated)}} \\*

& {\scriptsize hop2} &
{}Given the fields \textasciigrave{}question\textasciigrave{}, \textasciigrave{}summary\_\allowbreak{}1\textasciigrave{}, produce the fields \textasciigrave{}query\textasciigrave{}. \\*

& {\scriptsize hop3} &
{}Given the fields \textasciigrave{}question\textasciigrave{}, \textasciigrave{}context\textasciigrave{}, and \textasciigrave{}passages\textasciigrave{}, produce the field \textasciigrave{}summary\textasciigrave{}. Your task is to answer the question accurately and concisely, using the provided passages as the primary evidence source. The \textasciigrave{}context\textasciigrave{} may be helpful, but it can be incomplete or contain incorrect assumptions, so always verify important facts against the~{\itshape\textcolor{SelGray}{\ldots(truncated)}} \\*

& {\scriptsize hop4} &
{}Given the fields \textasciigrave{}question\textasciigrave{}, \textasciigrave{}summary\_\allowbreak{}1\textasciigrave{}, and \textasciigrave{}summary\_\allowbreak{}2\textasciigrave{}, produce the field \textasciigrave{}answer\textasciigrave{}. Your task is to answer the question as a concise factoid answer. Use the two summaries as evidence when they contain the needed information, but do not blindly trust summaries that say the information is unavailable, unknown, or that no connection exists. If the~{\itshape\textcolor{SelGray}{\ldots(truncated)}} \\
\addlinespace[2pt]
{}\textbf{Hotpot\allowbreak{}QA}\newline \textcolor{NaiveName}{\bfseries NPO}
& {\scriptsize hop1} &
{}You are the first-hop evidence summarization module in a multi-hop QA system. INPUT - question: the user\textquoteright{}s question - passages: a ranked list of retrieved passages, each with: - title - text OUTPUT FORMAT --- STRICT - Output exactly ONE valid JSON object. - The object must have exactly ONE key: \textquotedbl{}summary\textquotedbl{}. - Do not output any other keys, including~{\itshape\textcolor{SelGray}{\ldots(truncated)}} \\*

& {\scriptsize hop2} &
{}You are generating the SECOND-HOP retrieval query for a multi-hop QA system. INPUT FORMAT You will receive: - question: the original natural-language question. - summary\_\allowbreak{}1: a short summary of documents/\allowbreak{}evidence retrieved in the first hop. It may include correct evidence, wrong/\allowbreak{}ambiguous entities, partial answers, or statements such as~{\itshape\textcolor{SelGray}{\ldots(truncated)}} \\*

& {\scriptsize hop3} &
{}You are the summary generation module in a multi-hop QA system. INPUT FORMAT You will receive: - question: the final question to be answered. - context: an intermediate/\allowbreak{}previous summary. It may be useful, incomplete, misleading, or wrong. - passages: newly retrieved passages, each with a title and text. Passages may be relevant, partially~{\itshape\textcolor{SelGray}{\ldots(truncated)}} \\*

& {\scriptsize hop4} &
{}You are given three input fields: - question - summary\_\allowbreak{}1 - summary\_\allowbreak{}2 Produce exactly one output field: - answer Task: Answer the question with the most concise factual answer that satisfies the exact wording of the question. This is a multi-hop QA task: identify the entity or fact being asked about, connect it through the clues in the question~{\itshape\textcolor{SelGray}{\ldots(truncated)}} \\
\addlinespace[2pt]
{}\textbf{IFBench}\newline \textcolor{GEPAname}{\bfseries GEPA}
& {\scriptsize hop1} &
{}respond directly to the user\textquotesingle{}s query while satisfying every explicit requirement in the query. before answering, identify and obey all constraints in the prompt, including: - required language, casing, wording, keywords, exact phrases, section counts, headings, markdown formatting, and ending text. - if the user specifies exact labels or phrases~{\itshape\textcolor{SelGray}{\ldots(truncated)}} \\*

& {\scriptsize hop2} &
{}You are producing the final answer to the user\textquoteright{}s query. First solve the underlying task correctly, then format the final response so that every stated constraint is satisfied exactly. General process: 1. Parse the query for: - The actual task to solve, such as math, translation, natural language inference, or explanation. - Output-format~{\itshape\textcolor{SelGray}{\ldots(truncated)}} \\
\addlinespace[2pt]
{}\textbf{IFBench}\newline \textcolor{NaiveName}{\bfseries NPO}
& {\scriptsize hop1} &
{}You are an assistant whose job is to respond directly to the provided \textasciigrave{}query\textasciigrave{} while strictly satisfying every explicit instruction inside it. Core response rules: 1. Do not reveal hidden reasoning or internal chain-of-thought. Provide only the final answer requested by the user. 2. Carefully parse all formatting, length, word-count~{\itshape\textcolor{SelGray}{\ldots(truncated)}} \\*

& {\scriptsize hop2} &
{}You are given a task input containing at least: - query: the user\textquoteright{}s original request, including all formatting/\allowbreak{}content constraints - response: a prior assistant response that may be incomplete, incorrect, or may violate constraints Your job is to produce the corrected final answer that should be returned to the user. Do not merely evaluate the~{\itshape\textcolor{SelGray}{\ldots(truncated)}} \\
\addlinespace[2pt]
\multicolumn{2}{>{\raggedright\arraybackslash}p{0.20\textwidth}}{{}\textbf{2048}\newline \textcolor{GEPAname}{\bfseries GEPA}}
&
{}You play 2048 on a 3\ensuremath{\times}3 board. Output one move only. STRICT OUTPUT: - Final non-empty line must be exactly one of: [up] [down] [left] [right] - No coordinates, no multiple moves, no extra words on final line. - If legal/\allowbreak{}valid actions are listed, choose only a listed direction. - Prefer no reasoning; if reasoning, keep it to one short line before~{\itshape\textcolor{SelGray}{\ldots(truncated)}} \\
\addlinespace[2pt]
\multicolumn{2}{>{\raggedright\arraybackslash}p{0.20\textwidth}}{{}\textbf{2048}\newline \textcolor{NaiveName}{\bfseries NPO}}
&
{}You play 2048 on a 3\ensuremath{\times}3 board. OUTPUT FORMAT IS CRITICAL: Your entire response must be exactly one line and exactly one of: [up] [down] [left] [right] No \textless{}think\textgreater{}, no explanation, no coordinates, no punctuation, no extra text. Never output more than one move. If you previously made an invalid/\allowbreak{}no-change move on this same board, NEVER repeat that~{\itshape\textcolor{SelGray}{\ldots(truncated)}} \\
\addlinespace[2pt]
\multicolumn{2}{>{\raggedright\arraybackslash}p{0.20\textwidth}}{{}\textbf{Blackjack}\newline \textcolor{GEPAname}{\bfseries GEPA}}
&
{}You are Player 0 in Blackjack. Choose the action that maximizes win rate. Output exactly one line, with exactly one valid action: [Hit] or [Stand] No explanation, no extra text, no thinking tags. Use only the current hand and dealer upcard. Ignore past results. Hand type: - Soft = an Ace can be counted as 11 in the displayed score without busting.~{\itshape\textcolor{SelGray}{\ldots(truncated)}} \\
\addlinespace[2pt]
\multicolumn{2}{>{\raggedright\arraybackslash}p{0.20\textwidth}}{{}\textbf{Blackjack}\newline \textcolor{NaiveName}{\bfseries NPO}}
&
{}You are Player 0 in TextArena Blackjack. Maximize final reward across 20 hands. Dealer draws to 17. CRITICAL OUTPUT RULE: Your entire response must be exactly one line: [Hit] or [Stand] No reasoning, no \textless{}think\textgreater{}, no explanation, no extra text. IMPORTANT STRATEGY FOR THIS ENV: There is a 21-turn limit and unplayed hands score as zero. Each [Hit]~{\itshape\textcolor{SelGray}{\ldots(truncated)}} \\
\addlinespace[2pt]
\multicolumn{2}{>{\raggedright\arraybackslash}p{0.20\textwidth}}{{}\textbf{Briscola}\newline \textcolor{GEPAname}{\bfseries GEPA}}
&
{}You are Player 1 in 2-player Briscola. You see only your hand plus public info. Goal: maximize raw points captured out of 120. Rules: - Italian 40-card deck. Each player holds up to 3 cards and draws after each trick while deck lasts. - Points: A=11, 3=10, K=4, Q=3, J=2, 7/\allowbreak{}6/\allowbreak{}5/\allowbreak{}4/\allowbreak{}2=0. - Power within a suit: A \textgreater{} 3 \textgreater{} K \textgreater{} Q \textgreater{} J \textgreater{} 7 \textgreater{} 6 \textgreater{} 5 \textgreater{} 4 \textgreater{} 2.~{\itshape\textcolor{SelGray}{\ldots(truncated)}} \\
\addlinespace[2pt]
\multicolumn{2}{>{\raggedright\arraybackslash}p{0.20\textwidth}}{{}\textbf{Briscola}\newline \textcolor{NaiveName}{\bfseries NPO}}
&
{}You are Player 1 in 2-player Briscola. Maximize RAW POINTS captured out of 120; \textgreater{}60 wins, but keep taking points for the biggest margin. Rules: 40-card Italian deck, suits \ensuremath{\spadesuit}\ensuremath{\heartsuit}\ensuremath{\diamondsuit}\ensuremath{\clubsuit}, ranks A,2,3,4,5,6,7,J,Q,K. Up to 3 cards in hand; winner of each trick takes both cards\textquoteright{} points, leads next, and draws first while deck remains. Points: A=11, 3=10, K=4~{\itshape\textcolor{SelGray}{\ldots(truncated)}} \\
\addlinespace[2pt]
\multicolumn{2}{>{\raggedright\arraybackslash}p{0.20\textwidth}}{{}\textbf{Checkers}\newline \textcolor{GEPAname}{\bfseries GEPA}}
&
{}You are Player 1: Black. Your pieces are b/\allowbreak{}B (\ensuremath{\bullet}/\allowbreak{}\ensuremath{\blacksquare}); Red is r/\allowbreak{}R (\textcolor{red}{\ensuremath{\bullet}}). Move only Black. Black men move toward row 7 (increasing row). Black kings move both directions. A Black man reaching row 7 becomes a king. Use ONLY the current Valid Moves list. Do not invent moves from the board. Do not change, reorder, or \textquotedblleft{}fix\textquotedblright{} coordinates. Before answering~{\itshape\textcolor{SelGray}{\ldots(truncated)}} \\
\addlinespace[2pt]
\multicolumn{2}{>{\raggedright\arraybackslash}p{0.20\textwidth}}{{}\textbf{Checkers}\newline \textcolor{NaiveName}{\bfseries NPO}}
&
{}You are Player 1, Black. Black pieces are \ensuremath{\bullet}/\allowbreak{}\ensuremath{\blacksquare} (\textasciigrave{}b\textasciigrave{}/\allowbreak{}\textasciigrave{}B\textasciigrave{}); Red pieces are \textcolor{red}{\ensuremath{\bullet}}/\allowbreak{}\textcolor{red}{\ensuremath{\blacksquare}} (\textasciigrave{}r\textasciigrave{}/\allowbreak{}\textasciigrave{}R\textasciigrave{}). Move only Black. Black men move toward row 7; Black kings move both directions. Absolute legality: 1. Your final line must be exactly one bracketed move with four numbers, e.g. \textasciigrave{}[2 1 3 2]\textasciigrave{}. 2. Normally, copy exactly ONE move from \textasciigrave{}Valid Moves\textasciigrave{} verbatim. Do not~{\itshape\textcolor{SelGray}{\ldots(truncated)}} \\
\addlinespace[2pt]
\multicolumn{2}{>{\raggedright\arraybackslash}p{0.20\textwidth}}{{}\textbf{Chess}\newline \textcolor{GEPAname}{\bfseries GEPA}}
&
{}You are Player 1: Black. Use only one move copied exactly from the Valid Moves list. Final line must be exactly one bracketed UCI move, e.g. [g8f6]. No commentary. Do not get confused by board orientation or Player Id text: valid moves are legal. Choose the best legal move from the list. Selection priority: 1. If in check, get safe; prefer~{\itshape\textcolor{SelGray}{\ldots(truncated)}} \\
\addlinespace[2pt]
\multicolumn{2}{>{\raggedright\arraybackslash}p{0.20\textwidth}}{{}\textbf{Chess}\newline \textcolor{NaiveName}{\bfseries NPO}}
&
{}You are Player 1. Usually play Black (lowercase), but the observation is final authority: always move the \textquotedblleft{}Side to move\textquotedblright{} shown, using only moves from Valid Moves. OUTPUT --- absolute priority: - Output exactly one line only: one bracketed UCI move copied verbatim from Valid Moves, e.g. [e7e5]. - No analysis, no \textless{}think\textgreater{}, no explanation, no extra~{\itshape\textcolor{SelGray}{\ldots(truncated)}} \\
\addlinespace[2pt]
\multicolumn{2}{>{\raggedright\arraybackslash}p{0.20\textwidth}}{{}\textbf{Chopsticks}\newline \textcolor{GEPAname}{\bfseries GEPA}}
&
{}You are Player 1 in Chopsticks. Win by making Player 0 hands \textasciigrave{}[0,0]\textasciigrave{}. Rules: - Hands 0 and 1 hold 0-4 fingers; 0 is dead. - Attack \textasciigrave{}[attack M O]\textasciigrave{}: your live hand M adds its value to Player 0\textquoteright{}s live hand O. If target becomes 5 or more, it becomes 0. Your hand does not change. Never use a 0 hand; never target a 0 hand. - Split \textasciigrave{}[split L R]\textasciigrave{}~{\itshape\textcolor{SelGray}{\ldots(truncated)}} \\
\addlinespace[2pt]
\multicolumn{2}{>{\raggedright\arraybackslash}p{0.20\textwidth}}{{}\textbf{Chopsticks}\newline \textcolor{NaiveName}{\bfseries NPO}}
&
{}You are Player 1 in Chopsticks. Opponent is Player 0. Win by making Player 0 hands [0,0]. Always output one LEGAL move. Rules: - Hands are indexed 0 and 1. Values 1-4 are live; 0 is dead. - Attack \textasciigrave{}[attack M O]\textasciigrave{}: my live hand M adds its value to Player 0\textquoteright{}s live target O. If target becomes 5 or more, it becomes 0. My attacking hand does not change.~{\itshape\textcolor{SelGray}{\ldots(truncated)}} \\
\addlinespace[2pt]
\multicolumn{2}{>{\raggedright\arraybackslash}p{0.20\textwidth}}{{}\textbf{Connect\allowbreak{}Four}\newline \textcolor{GEPAname}{\bfseries GEPA}}
&
{}You are Player 1 in Connect Four. Your disc is O. Opponent/\allowbreak{}Player 0 is X. Board/\allowbreak{}rules: - Board is 6 rows \ensuremath{\times} 7 columns. - Columns are indexed left to right: 0 1 2 3 4 5 6. - A move chooses exactly one column; the disc falls to the lowest empty cell in that column. - Win by connecting four of your O discs vertically, horizontally, or diagonally.~{\itshape\textcolor{SelGray}{\ldots(truncated)}} \\
\addlinespace[2pt]
\multicolumn{2}{>{\raggedright\arraybackslash}p{0.20\textwidth}}{{}\textbf{Connect\allowbreak{}Four}\newline \textcolor{NaiveName}{\bfseries NPO}}
&
{}You are Player 1 in Connect Four. Your disc is O. Opponent is X. Board/\allowbreak{}rules: - 6 rows \ensuremath{\times} 7 columns. - Choose exactly one column: 0,1,2,3,4,5,6. - A disc falls to the lowest empty cell in that column. - Win by connecting four O discs vertically, horizontally, or diagonally. - First identify O as yours and X as opponent. Legal move rule is~{\itshape\textcolor{SelGray}{\ldots(truncated)}} \\
\addlinespace[2pt]
\multicolumn{2}{>{\raggedright\arraybackslash}p{0.20\textwidth}}{{}\textbf{Countdown}\newline \textcolor{GEPAname}{\bfseries GEPA}}
&
{}You are Player 0 in single-player Countdown. Each turn, read the CURRENT TARGET and CURRENT Available numbers list, then output one legal move. MOVE FORMAT /\allowbreak{} LEGALITY - A move is exactly \textasciigrave{}[i j op]\textasciigrave{}. - \textasciigrave{}i\textasciigrave{} and \textasciigrave{}j\textasciigrave{} are 0-based INDICES into the CURRENT list only. Re-read the list every turn because indices change after each move. - If there are n~{\itshape\textcolor{SelGray}{\ldots(truncated)}} \\
\addlinespace[2pt]
\multicolumn{2}{>{\raggedright\arraybackslash}p{0.20\textwidth}}{{}\textbf{Countdown}\newline \textcolor{NaiveName}{\bfseries NPO}}
&
{}You are Player 0 in single-player Countdown. Each turn choose exactly ONE legal move that helps hit the TARGET or get closest. LEGALITY IS FIRST - Re-read the CURRENT Available numbers list every turn. The list shrinks by one after each move and indices change. - A move is exactly \textasciigrave{}[i j op]\textasciigrave{}: \textasciigrave{}i\textasciigrave{} and \textasciigrave{}j\textasciigrave{} are 0-based indices in the CURRENT list, \textasciigrave{}i~{\itshape\textcolor{SelGray}{\ldots(truncated)}} \\
\addlinespace[2pt]
\multicolumn{2}{>{\raggedright\arraybackslash}p{0.20\textwidth}}{{}\textbf{Crusade}\newline \textcolor{GEPAname}{\bfseries GEPA}}
&
{}You are Player 1 in Crusade, playing Black (\textquotesingle{}B\textquotesingle{}). Opponent is White (\textquotesingle{}W\textquotesingle{}); empty squares are \textquotesingle{}.\textquotesingle{}. Use only the current board and the current \textquotedbl{}Available Moves\textquotedbl{} list. Rules: every piece moves like a chess knight (L-shape). A legal move is \textasciigrave{}[from to]\textasciigrave{}. You may move a Black piece to an empty square or onto a White piece; landing on \textasciigrave{}W\textasciigrave{} captures it for~{\itshape\textcolor{SelGray}{\ldots(truncated)}} \\
\addlinespace[2pt]
\multicolumn{2}{>{\raggedright\arraybackslash}p{0.20\textwidth}}{{}\textbf{Crusade}\newline \textcolor{NaiveName}{\bfseries NPO}}
&
{}You are Player 1 in Crusade. You play Black (\textasciigrave{}B\textasciigrave{}). White (\textasciigrave{}W\textasciigrave{}) is the opponent. Empty squares are \textasciigrave{}.\textasciigrave{}. OUTPUT RULES: - Use only the current Board and current Available Moves. - The final non-empty line must be exactly one move copied verbatim from Available Moves, like \textasciigrave{}[g8 f6]\textasciigrave{}. - Nothing after the final move line. - Never output a move not in~{\itshape\textcolor{SelGray}{\ldots(truncated)}} \\
\addlinespace[2pt]
\multicolumn{2}{>{\raggedright\arraybackslash}p{0.20\textwidth}}{{}\textbf{Kuhn\allowbreak{}Poker}\newline \textcolor{GEPAname}{\bfseries GEPA}}
&
{}You are Player 1 in 2-player Kuhn Poker against fixed Player 0. Each round both players ante 1 chip, receive one private card from \{J,Q,K\} with J \textless{} Q \textless{} K, then check/\allowbreak{}bet/\allowbreak{}call/\allowbreak{}fold. You see only your own card. If both check or someone calls, higher card wins at showdown; if someone folds, the other wins the pot. Net chips accumulate across rounds.~{\itshape\textcolor{SelGray}{\ldots(truncated)}} \\
\addlinespace[2pt]
\multicolumn{2}{>{\raggedright\arraybackslash}p{0.20\textwidth}}{{}\textbf{Kuhn\allowbreak{}Poker}\newline \textcolor{NaiveName}{\bfseries NPO}}
&
{}You are Player 1 in 2-player Kuhn Poker. Maximize Player 1 raw net chips across all rounds. CRITICAL OUTPUT RULE: Output ONE LINE ONLY: exactly one lowercase bracketed action: [check], [bet], [call], or [fold]. Do not write reasoning, \textless{}think\textgreater{}, explanations, or extra text. The single line you output must be the action you intend. Game facts: each~{\itshape\textcolor{SelGray}{\ldots(truncated)}} \\
\addlinespace[2pt]
\multicolumn{2}{>{\raggedright\arraybackslash}p{0.20\textwidth}}{{}\textbf{Lights\allowbreak{}Out}\newline \textcolor{GEPAname}{\bfseries GEPA}}
&
{}You are Player 0 in single-player 5x5 Lights Out. Goal: make every light OFF (\textasciigrave{}.\textasciigrave{}). Rules: - Press \textasciigrave{}[row col]\textasciigrave{} toggles that cell plus orthogonal neighbors only. - Rows/\allowbreak{}cols are 0-indexed integers 0-4. - Never repeat the same press unless the board changed and it is clearly useful. Move policy: 1. Parse ONLY the current displayed board. Do not use~{\itshape\textcolor{SelGray}{\ldots(truncated)}} \\
\addlinespace[2pt]
\multicolumn{2}{>{\raggedright\arraybackslash}p{0.20\textwidth}}{{}\textbf{Lights\allowbreak{}Out}\newline \textcolor{NaiveName}{\bfseries NPO}}
&
{}You are Player 0 playing single-player 5x5 Lights Out. Goal: make every cell \textasciigrave{}.\textasciigrave{}. Rules: - A press \textasciigrave{}[r c]\textasciigrave{} toggles that cell and its orthogonal neighbors only: up/\allowbreak{}down/\allowbreak{}left/\allowbreak{}right. - Rows and columns are 0-indexed integers 0-4. - Pressing the same cell twice cancels out. Do not repeat a coordinate already pressed in this episode unless it is part~{\itshape\textcolor{SelGray}{\ldots(truncated)}} \\
\addlinespace[2pt]
\multicolumn{2}{>{\raggedright\arraybackslash}p{0.20\textwidth}}{{}\textbf{Logic\allowbreak{}Puzzle}\newline \textcolor{GEPAname}{\bfseries GEPA}}
&
{}You are Player 0 solving a TextArena logic-grid puzzle. Make exactly one forced mark per turn. Critical board-reading rules: - Row labels are on the LEFT. Column labels are at the TOP. Ignore the grid title except to know which two categories are paired. - Your action must use \textasciigrave{}[row col O]\textasciigrave{} or \textasciigrave{}[row col X]\textasciigrave{} with the row label first and column~{\itshape\textcolor{SelGray}{\ldots(truncated)}} \\
\addlinespace[2pt]
\multicolumn{2}{>{\raggedright\arraybackslash}p{0.20\textwidth}}{{}\textbf{Logic\allowbreak{}Puzzle}\newline \textcolor{NaiveName}{\bfseries NPO}}
&
{}You solve ONE turn of a TextArena logic-grid puzzle. Mark exactly ONE forced, visibly BLANK cell. Never guess. MANDATORY FINAL ACTION: - The final non-empty line must be exactly \textasciigrave{}[row col O]\textasciigrave{} or \textasciigrave{}[row col X]\textasciigrave{} - Use row label from the LEFT and column label from the TOP exactly as printed. - \textasciigrave{}O\textasciigrave{} = this pair is true. \textasciigrave{}X\textasciigrave{} = this pair is false.~{\itshape\textcolor{SelGray}{\ldots(truncated)}} \\
\addlinespace[2pt]
\multicolumn{2}{>{\raggedright\arraybackslash}p{0.20\textwidth}}{{}\textbf{Minesweeper}\newline \textcolor{GEPAname}{\bfseries GEPA}}
&
{}You are Player 0 playing single-player Minesweeper. Reveal one safe unrevealed cell and avoid mines. Rules: - Coordinates are 0-indexed row col from the shown axes. - \textquotedbl{}.\textquotedbl{} means unrevealed and is the ONLY legal target. - Numbers are already revealed safe cells; NEVER choose them. - Minesweeper adjacency is all 8 surrounding cells, including~{\itshape\textcolor{SelGray}{\ldots(truncated)}} \\
\addlinespace[2pt]
\multicolumn{2}{>{\raggedright\arraybackslash}p{0.20\textwidth}}{{}\textbf{Minesweeper}\newline \textcolor{NaiveName}{\bfseries NPO}}
&
{}You are Player 0 playing single-player Minesweeper. Think internally, but OUTPUT EXACTLY 4 LINES ONLY. No \textasciigrave{}\textless{}think\textgreater{}\textasciigrave{}, no markdown, no extra text. Line 1: Known safe: coordinates or none Line 2: Known mines: coordinates or none Line 3: Move reason: brief Line 4: [row col] STRICT FORMAT: - Line 4 must be exactly one coordinate copied from a current~{\itshape\textcolor{SelGray}{\ldots(truncated)}} \\
\addlinespace[2pt]
\multicolumn{2}{>{\raggedright\arraybackslash}p{0.20\textwidth}}{{}\textbf{Nim}\newline \textcolor{GEPAname}{\bfseries GEPA}}
&
{}You are a Nim agent. You are player 1; opponent is player 0. Count objects by counting \ensuremath{\bullet} in each row; empty rows have 0. A move removes at least 1 object from exactly ONE nonempty row. Taking the last object means all rows become empty. Strategy: 1. If only one row has objects, take all of it and win. 2. Compute nim-sum XOR of all row counts. 3.~{\itshape\textcolor{SelGray}{\ldots(truncated)}} \\
\addlinespace[2pt]
\multicolumn{2}{>{\raggedright\arraybackslash}p{0.20\textwidth}}{{}\textbf{Nim}\newline \textcolor{NaiveName}{\bfseries NPO}}
&
{}You are Player 1 in normal-play Nim. Each Row is one pile. Pile size is ONLY the count of \ensuremath{\bullet} symbols in that row; empty cells are 0. Output format must be exactly: [pile\_\allowbreak{}index quantity\_\allowbreak{}removed] The second number is the number to remove, NOT the target pile size. Move policy: 1. Count all pile sizes from \ensuremath{\bullet}. 2. If exactly one pile is nonempty, remove~{\itshape\textcolor{SelGray}{\ldots(truncated)}} \\
\addlinespace[2pt]
\multicolumn{2}{>{\raggedright\arraybackslash}p{0.20\textwidth}}{{}\textbf{Othello}\newline \textcolor{GEPAname}{\bfseries GEPA}}
&
{}You are Player 1 (White) on a 4x4 Othello board, rows/\allowbreak{}cols 0-3. Opponent is Player 0 (Black). Win by final disc count. Use ONLY the listed Valid Moves. Never invent a move, never play occupied cells, never play outside the list. If only one valid move, play it. Think briefly before moving: 1. If a legal corner [0 0], [0 3], [3 0], or [3 3] is~{\itshape\textcolor{SelGray}{\ldots(truncated)}} \\
\addlinespace[2pt]
\multicolumn{2}{>{\raggedright\arraybackslash}p{0.20\textwidth}}{{}\textbf{Othello}\newline \textcolor{NaiveName}{\bfseries NPO}}
&
{}You are Player 1, White, in 4x4 Othello. Rows/\allowbreak{}cols are 0-3. Black is Player 0. Win by final disc count. The shown Valid Moves list is authoritative. Choose exactly one coordinate from it. Do NOT reject or second-guess a listed valid move. Never play outside Valid Moves. Think briefly, then choose by this order: 1. If only one Valid Move, play it.~{\itshape\textcolor{SelGray}{\ldots(truncated)}} \\
\addlinespace[2pt]
\multicolumn{2}{>{\raggedright\arraybackslash}p{0.20\textwidth}}{{}\textbf{Pig\allowbreak{}Dice}\newline \textcolor{GEPAname}{\bfseries GEPA}}
&
{}You are Player 1 in Pig. Use only these fields: Scores [Player0, Player1], Turn Total, Turn, Winning Score. Ignore Board/\allowbreak{}Current roll/\allowbreak{}Goal if inconsistent. Your entire response must be exactly one line: [roll] or [hold]. No explanation. No \textless{}think\textgreater{}. Policy: 1. If Turn Total = 0: [roll]. 2. If Player1 score + Turn Total \textgreater{}= Winning Score: [hold]. 3.~{\itshape\textcolor{SelGray}{\ldots(truncated)}} \\
\addlinespace[2pt]
\multicolumn{2}{>{\raggedright\arraybackslash}p{0.20\textwidth}}{{}\textbf{Pig\allowbreak{}Dice}\newline \textcolor{NaiveName}{\bfseries NPO}}
&
{}You are Player 1 in Pig. Read only: Scores [Player0, Player1], Turn Total T, Turn number, and board Goal = 100. Ignore \textquotedblleft{}Winning Score\textquotedblright{}, Current roll, and Turn Rolls. OUTPUT RULE: Your entire response must be exactly one line: [roll] or [hold] No \textless{}think\textgreater{}, no explanation, no extra text. Let my = Player1 score, opp = Player0 score, lead = my - opp.~{\itshape\textcolor{SelGray}{\ldots(truncated)}} \\
\addlinespace[2pt]
\multicolumn{2}{>{\raggedright\arraybackslash}p{0.20\textwidth}}{{}\textbf{Quantum\allowbreak{}Tic\allowbreak{}Tac\allowbreak{}Toe}\newline \textcolor{GEPAname}{\bfseries GEPA}}
&
{}You are Player 1 (X). Your move numbers are ODD: X1, X3, X5... Opponent is Player 0 (O), EVEN moves. Trust \textquotedblleft{}Current Player: 1\textquotedblright{} even if the printed turn number is confusing. Board: 0 1 2 3 4 5 6 7 8 Winning lines: 012, 345, 678, 036, 147, 258, 048, 246. Win by getting 3 CLASSICAL solid X marks in a line. If both players complete a line in one~{\itshape\textcolor{SelGray}{\ldots(truncated)}} \\
\addlinespace[2pt]
\multicolumn{2}{>{\raggedright\arraybackslash}p{0.20\textwidth}}{{}\textbf{Quantum\allowbreak{}Tic\allowbreak{}Tac\allowbreak{}Toe}\newline \textcolor{NaiveName}{\bfseries NPO}}
&
{}You are Player 1 in Quantum Tic Tac Toe. You are X and your move numbers are odd (X1, X3, X5...). Act only when Current Player is 1. Board cells: 0 1 2 3 4 5 6 7 8 LEGALITY FIRST: - Your final non-empty line must be exactly \textasciigrave{}[a,b]\textasciigrave{} with no extra text. - \textasciigrave{}a\textasciigrave{} and \textasciigrave{}b\textasciigrave{} must be two DIFFERENT numbers from 0-8. - The exact pair \textasciigrave{}[a,b]\textasciigrave{} must appear in~{\itshape\textcolor{SelGray}{\ldots(truncated)}} \\
\addlinespace[2pt]
\multicolumn{2}{>{\raggedright\arraybackslash}p{0.20\textwidth}}{{}\textbf{Set}\newline \textcolor{GEPAname}{\bfseries GEPA}}
&
{}You are Player 0 playing single-player Set. Goal: maximize correct Sets. Wrong Sets waste turns. Format errors are forbidden. Card attributes are exactly: - number: one, two, three - color: red, green, purple - fill: open, solid, striped - shape: diamond, oval, squiggle A valid Set is exactly 3 cards where, for EACH attribute, the 3 values are all~{\itshape\textcolor{SelGray}{\ldots(truncated)}} \\
\addlinespace[2pt]
\multicolumn{2}{>{\raggedright\arraybackslash}p{0.20\textwidth}}{{}\textbf{Set}\newline \textcolor{NaiveName}{\bfseries NPO}}
&
{}You are Player 0 playing single-player Set. Choose exactly ONE valid Set from the CURRENT board. Set rule: for EACH attribute (number, color, fill, shape), the 3 cards must have either 1 distinct value (all same) or 3 distinct values (all different). If any attribute has exactly 2 distinct values, the trio is invalid. Strict procedure: 1. Parse~{\itshape\textcolor{SelGray}{\ldots(truncated)}} \\
\addlinespace[2pt]
\multicolumn{2}{>{\raggedright\arraybackslash}p{0.20\textwidth}}{{}\textbf{Simple\allowbreak{}Tak}\newline \textcolor{GEPAname}{\bfseries GEPA}}
&
{}You are Player 1 (X) in SimpleTak on a 4x4 board: 0 1 2 3 4 5 6 7 8 9 10 11 12 13 14 15 Win with an orthogonally connected X-chain touching TOP+BOTTOM or LEFT+RIGHT. Diagonals never connect. Critical rules: - Choose ONLY a cell listed in Valid Moves. - Trust the board rows above: below 3 is 7, below 7 is 11, below 11 is 15. - Before strategy~{\itshape\textcolor{SelGray}{\ldots(truncated)}} \\
\addlinespace[2pt]
\multicolumn{2}{>{\raggedright\arraybackslash}p{0.20\textwidth}}{{}\textbf{Simple\allowbreak{}Tak}\newline \textcolor{NaiveName}{\bfseries NPO}}
&
{}You are Player 1 (X) in SimpleTak. Output exactly one legal move. LAST line must be only \textasciigrave{}[cell]\textasciigrave{}. No explanation. Use Valid Moves as the legality source: choose only a bracketed cell listed there. Board indices: 0 1 2 3 4 5 6 7 8 9 10 11 12 13 14 15 Edges: Top=\{0,1,2,3\}; Bottom=\{12,13,14,15\}; Left=\{0,4,8,12\}; Right=\{3,7,11,15\}. Rows: \{0,1,2,3\}~{\itshape\textcolor{SelGray}{\ldots(truncated)}} \\
\addlinespace[2pt]
\multicolumn{2}{>{\raggedright\arraybackslash}p{0.20\textwidth}}{{}\textbf{Sokoban}\newline \textcolor{GEPAname}{\bfseries GEPA}}
&
{}You are Player 0 playing Sokoban. Your entire response must be exactly one line: [up], [down], [left], or [right]. No reasoning text. Before outputting, silently do this checklist: 1. Parse ONLY the Current Board. Rows/\allowbreak{}columns are the space-separated symbols. Locate P exactly. 2. Symbols: \# wall, \_\allowbreak{} floor, O target, X box, \ensuremath{\surd} box on target. 3. For~{\itshape\textcolor{SelGray}{\ldots(truncated)}} \\
\addlinespace[2pt]
\multicolumn{2}{>{\raggedright\arraybackslash}p{0.20\textwidth}}{{}\textbf{Sokoban}\newline \textcolor{NaiveName}{\bfseries NPO}}
&
{}You are Player 0 playing Sokoban. Reply with exactly ONE legal move and nothing else: \textasciigrave{}[up]\textasciigrave{}, \textasciigrave{}[down]\textasciigrave{}, \textasciigrave{}[left]\textasciigrave{}, or \textasciigrave{}[right]\textasciigrave{}. No reasoning, no \textasciigrave{}\textless{}think\textgreater{}\textasciigrave{}, no extra text. Use the latest \textasciigrave{}Current Board:\textasciigrave{}. Symbols: \textasciigrave{}\#\textasciigrave{} wall, \textasciigrave{}\_\allowbreak{}\textasciigrave{} floor, \textasciigrave{}P\textasciigrave{} player, \textasciigrave{}X\textasciigrave{} box, \textasciigrave{}O\textasciigrave{} target, \textasciigrave{}\ensuremath{\surd}\textasciigrave{} box on target. Critical invalid-retry rule: - If the observation says~{\itshape\textcolor{SelGray}{\ldots(truncated)}} \\
\addlinespace[2pt]
\multicolumn{2}{>{\raggedright\arraybackslash}p{0.20\textwidth}}{{}\textbf{Wild\allowbreak{}Tic\allowbreak{}Tac\allowbreak{}Toe}\newline \textcolor{GEPAname}{\bfseries GEPA}}
&
{}You are Player 1 in Wild Tic Tac Toe. Board cells: 0 1 2 3 4 5 6 7 8 Lines: 012, 345, 678, 036, 147, 258, 048, 246. Rules: On every turn you choose BOTH a mark (X or O) and an empty cell. Either player may place either mark. Whoever completes any line of three identical marks wins immediately, no matter who placed the earlier marks. Critical~{\itshape\textcolor{SelGray}{\ldots(truncated)}} \\
\addlinespace[2pt]
\multicolumn{2}{>{\raggedright\arraybackslash}p{0.20\textwidth}}{{}\textbf{Wild\allowbreak{}Tic\allowbreak{}Tac\allowbreak{}Toe}\newline \textcolor{NaiveName}{\bfseries NPO}}
&
{}You are Player 1 in Wild Tic Tac Toe. Board cells: 0 1 2 3 4 5 6 7 8 Winning lines: 012, 345, 678, 036, 147, 258, 048, 246. Rules: - On every turn choose BOTH a mark (\textasciigrave{}X\textasciigrave{} or \textasciigrave{}O\textasciigrave{}) and an EMPTY cell. - Either player may place either mark. Marks are shared. - Whoever places the third identical mark in any line wins immediately: \textasciigrave{}XXX\textasciigrave{} or \textasciigrave{}OOO\textasciigrave{}, no~{\itshape\textcolor{SelGray}{\ldots(truncated)}} \\
\addlinespace[2pt]
\multicolumn{2}{>{\raggedright\arraybackslash}p{0.20\textwidth}}{{}\textbf{Wordle}\newline \textcolor{GEPAname}{\bfseries GEPA}}
&
{}You are Player 0 playing single-player Wordle. Each turn output one valid guess only. Think privately and very briefly, then output only the action line. Do NOT write \textasciigrave{}\textless{}think\textgreater{}\textasciigrave{}, explanations, notes, or invented filler. Never output nonwords such as resub/\allowbreak{}respc/\allowbreak{}rescd/\allowbreak{}rescg/\allowbreak{}ixsos. Use all board feedback together: - G: keep that letter fixed in that~{\itshape\textcolor{SelGray}{\ldots(truncated)}} \\
\addlinespace[2pt]
\multicolumn{2}{>{\raggedright\arraybackslash}p{0.20\textwidth}}{{}\textbf{Wordle}\newline \textcolor{NaiveName}{\bfseries NPO}}
&
{}You are Player 0 playing single-player Wordle. Make exactly one valid 5-letter English word guess. CRITICAL OUTPUT RULES: - Keep reasoning very brief or silent; do NOT write long analysis. - The LAST non-empty line must be exactly \textasciigrave{}[word]\textasciigrave{}. - \textasciigrave{}word\textasciigrave{} must be one real common English word, exactly 5 lowercase letters, not previously guessed. - Never~{\itshape\textcolor{SelGray}{\ldots(truncated)}} \\
\addlinespace[2pt]
\end{longtable}
\endgroup

\end{document}